\documentclass{article}
\usepackage[english]{babel}
\usepackage[utf8]{inputenc}
\usepackage{johd}
\usepackage{graphicx}
\usepackage{amsmath}
\usepackage{amssymb}
\usepackage{amsthm}
\usepackage{bm}
\usepackage{algorithm}
\usepackage{algorithmicx}
\usepackage{booktabs}
\usepackage{subfigure}
\usepackage[noend]{algpseudocode}
\usepackage[numbers]{natbib}
\newtheorem{theorem}{Theorem}

\newtheorem{lemma}{Lemma}

\newtheorem{definition}{Definition}

\title{Revisiting Counterfactual Regression through the Lens of Gromov-Wasserstein Information Bottleneck}

\author{Hao Yang, Zexu Sun, Hongteng Xu$^*$, Xu Chen$^*$ \\
        \small Gaoling School of Artificial Intelligence, Renmin University of China \\\\
        \small $^{*}$Corresponding authors: Hongteng Xu, Xu Chen\\\small  \tt{\{hongtengxu, xu.chen\}@ruc.edu.cn} \\
}
\date{}

\begin{document}
\maketitle
\begin{abstract} 

As a promising individualized treatment effect (ITE) estimation method, counterfactual regression (CFR) maps individuals' covariates to a latent space and predicts their counterfactual outcomes. 
However, the selection bias between control and treatment groups often imbalances the two groups' latent distributions and negatively impacts this method's performance.
In this study, we revisit counterfactual regression through the lens of information bottleneck and propose a novel learning paradigm called Gromov-Wasserstein information bottleneck (GWIB).
In this paradigm, we learn CFR by maximizing the mutual information between covariates' latent representations and outcomes while penalizing the kernelized mutual information between the latent representations and the covariates.
We demonstrate that the upper bound of the penalty term can be implemented as a new regularizer consisting of $i)$ the fused Gromov-Wasserstein distance between the latent representations of different groups and $ii)$ the gap between the transport cost generated by the model and the cross-group Gromov-Wasserstein distance between the latent representations and the covariates. 
GWIB effectively learns the CFR model through alternating optimization, suppressing selection bias while avoiding trivial latent distributions. 
Experiments on ITE estimation tasks show that GWIB consistently outperforms state-of-the-art CFR methods.
To promote the research community, we release our project at https://github.com/peteryang1031/Causal-GWIB.
\end{abstract}

\noindent\\

\section{Introduction}
\label{sec:intro}
Individualized treatment effect (ITE), which is usually estimated through conditional average treatment effect (CATE) estimation~\cite{abrevaya2015estimating} in real-world applications, plays a central role in causal inference. 
With the increasing availability and scale of observational data, estimating ITE based on counterfactual regression (CFR)~\cite{shalit2017estimating} rather than expensive randomized controlled trials (RCTs)~\cite{chalmers1981method} becomes a promising solution to this problem, which has been widely used in fields like healthcare~\cite{gueyffier1997effect}, economics~\cite{sun2023robustness}, and education~\cite{roberts2011using}.
As a price for low-cost utilization of massive data, counterfactual regression may suffer from the \textit{selection bias} between control and treatment groups~\cite{hernan2004structural}.
In particular, treatments are often not randomly assigned to individuals in practice, making the two groups' covariate distributions imbalanced.
Therefore, the counterfactual regression without careful adjustment may result in biased ITE estimation~\cite{yao2021survey, assaad2021counterfactual, wang2023optimal}.

To mitigate the selection bias, various regularization methods are applied to counterfactual regression~\cite{yao2021survey}.
In principle, counterfactual regression maps covariates to a latent space, and these methods apply various domain adaptation techniques~\cite{farahani2021brief} to balance the latent distributions of control and treatment groups.
Typical balancing strategies include $i)$ minimizing the discrepancy between the latent distributions (e.g., CFRNet~\cite{shalit2017estimating} and ESCFR~\cite{wang2023optimal}) and $ii)$ resampling and smoothing the latent distributions (e.g., StableCFR~\cite{wu2023stable}). 
However, the CFR model learned with inappropriate regularization tends to ignore the information of original covariate distributions and generates balanced but trivial latent distributions, giving rise to the so-called \textit{over-enforcing balance} issue~\cite{zhao2019learning, johansson2018learning}, which may inadvertently remove information predictive of outcomes, thus leading to undesired counterfactual predictions as well.

\begin{figure*}[t]
\centering
\includegraphics[width=\linewidth]{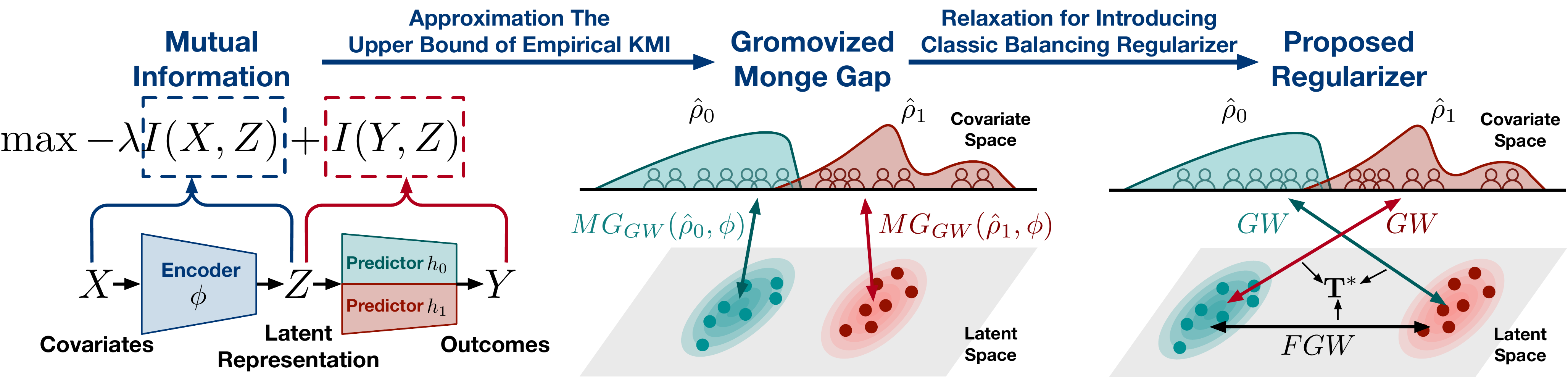}
\caption{The scheme of GWIB. 
$\hat{\rho}_0$ and $\hat{\rho}_1$ denote the empirical covariate distributions of control and treatment groups, respectively.
$\phi$ is the encoder mapping covariates to their latent representations. 
An upper bound of the empirical KMI is approximated based on the Gromovized Monge gap associated with $\phi$. 
Further considering the classic balancing penalty leads to the proposed regularizer, which involves the GW and FGW distances that share the same optimal transport plan $\bm{T}^*$.}\label{fig-motivation}
\end{figure*}

From the viewpoint of information bottleneck~\cite{tishby1999information}, both selection bias and over-enforcing balance are due to the failure to achieve a trade-off between accuracy and complexity (or, equivalently, compression) when representing covariates.
Ideally, the latent representations of covariates should only preserve necessary information for ITE estimation while filtering out useless information.
Therefore, in this study, we revisit CFR through the lens of information bottleneck and propose a novel Gromov-Wasserstein information bottleneck (GWIB) learning paradigm. 
This paradigm achieves a new and effective optimal transport-based solution to mitigate selection bias and over-enforcing balance issues jointly. 
As illustrated in Figure~\ref{fig-motivation}, we formulate the CFR learning problem as maximizing the mutual information between covariates' latent representations and outcomes while penalizing the kernelized mutual information~\cite{chuang2023infoot} between the latent representations and the covariates. 
As the main theoretical contribution of this work, we demonstrate that the upper bound of the penalty term corresponds to the Gromovized Monge gap~\cite{uscidda2023monge} associated with the encoder of CFR.
Furthermore, we can introduce the classic balancing strategy by relaxing the upper bound, which leads to the proposed regularizer consisting of $i)$ the fused Gromov-Wasserstein (FGW) distance~\cite{titouan2019optimal} between the latent representations of different groups and $ii)$ the gap between the transport cost generated by the model and the cross-group Gromov-Wasserstein distance~\cite{memoli2011gromov} between the latent representations and the covariates.
The FGW and GW distances share the same optimal transport plan (i.e., the $\bm{T}$ in Figure~\ref{fig-motivation}), leading to consistent cross-group individual correspondence inference. 

To our knowledge, our work makes the first attempt to build a Gromov-Wasserstein distance-based information bottleneck method, which connects the estimation of kernelized mutual information with the computation of Gromovized Monge gap.
The GWIB paradigm provides an insightful perspective to model CFR tasks --- regularizing the encoder to approximate the Monge map helps penalize the mutual information between covariates and their latent representations. 
In the GWIB paradigm, we can learn the CFR model and the OT plan effectively in an alternating optimization framework.
We conduct comprehensive experiments to verify the effectiveness of GWIB, and the results demonstrate that GWIB outperforms the state-of-the-art CFR methods in various ITE estimation tasks.

\section{Preliminaries}\label{preli-causal}
Let $x\in\mathcal{X}$ be the covariate associated with an individual, where $\mathcal{X}$ denotes a covariate space.
We denote $t\in\{0,1\}$ as the treatment determining whether a treatment is applied (i.e., $t=1$) or not (i.e., $t=0$). 
For the individual whose covariate is $x$, its factual outcome under the treatment $t$ is denoted as $y(x,t)\in\mathbb{R}$. 
Accordingly, its counterfactual outcome is denoted as $y(x,1-t)$, which does not happen in practice and thus is unavailable.
The treatment group consists of the individuals applying a treatment, denoted as $\{(x,y(x,1))|x\sim\rho_1\}$, in which the covariates of the individuals obey a distribution $\rho_1$. 
Similarly, the control group corresponding to the individuals not applying the treatment is denoted as $\{(x,y(x,0))|x\sim\rho_0\}$, in which the covariates obey a distribution $\rho_0$. 
In the classic potential outcome framework~\cite{rubin1974estimating,splawa1990application}, the ITE on an individual is defined as
\begin{eqnarray}\label{eq:ite}
    \begin{aligned}
        \tau(x) = y(x,1) - y(x,0),
    \end{aligned}
\end{eqnarray}
where either $y(x,1)$ or $y(x,0)$ is counterfactual and needs to be estimated. 
Counterfactual regression~\cite{shalit2017estimating} model estimate the ITE as 
\begin{eqnarray}\label{eq:pred_ite}
    \begin{aligned}
        \hat{\tau}(x) = \hat{y}(x,1) - \hat{y}(x,0) = h_1\circ \phi(x) - h_0\circ\phi(x),
    \end{aligned}
\end{eqnarray}
where ``$\circ$'' denotes the composition of two functions. $\phi:\mathcal{X}\mapsto\mathcal{Z}$ is the encoder mapping the covariates to a latent space $\mathcal{Z}$, i.e., $z=\phi(x)\in\mathcal{Z}$ for $x\in\mathcal{X}$.
Through $\phi$, we push the covariate distributions forward to two latent distributions, denoted as $\phi_{\#}\rho_0$ and $\phi_{\#}\rho_1$, respectively.
$h_0, h_1: \mathcal{Z} \mapsto \mathbb{R}$ are two prediction functions that take the latent representations as input and predict the potential outcomes for the control and treatment groups, respectively.

As shown in Theorem~1 in~\cite{shalit2017estimating}, 
the expected precision in estimation of heterogeneous effect~\cite{hill2011bayesian} (a key measurement of ITE estimation), i.e., $\epsilon_{PEHE}=\mathbb{E}_{x}[(\hat{\tau}(x)-\tau(x))^2]$, is bounded by $2\sum_{t=0,1}\epsilon_{t}(h_t,\phi)+ B_{\phi}IPM(\phi_{\#}\rho_0,\phi_{\#}\rho_1)-2\sigma_Y^2$.
Here, $\{\epsilon_t(h_t,\phi)\}_{t=0,1}$ are squared factual losses of control ($t=0$) and treatment ($t=1$) groups, respectively. $IPM(\phi_{\#}\rho_0,\phi_{\#}\rho_1)$ is the integral probability metric between the latent distributions of the two groups. 
$B_{\phi}>0$ is a coefficient determined by the encoder $\phi$.
$\sigma_Y^2$ denotes the variance of outcomes.
Based on this generalization error bound, given observed covariates and corresponding outcomes, i.e., $\mathcal{D}_t=\{x_{n,t}, y_{n,t})\}_{n=1}^{N_t}$, $t=0,1$, we often learn a CFR model by solving the following optimization problem:
\begin{eqnarray}\label{eq:classic_cfr}
    \sideset{}{_{h_0,h_1,\phi}}\min \underbrace{\sideset{}{_{t=0,1}}\sum\frac{1}{N_t}\sideset{}{_{n=1}^{N_t}}\sum|h_t\circ \phi(x_{n,t})-y_{n,t}|^2}_{\mathcal{L}(h_0,h_1,\phi)} + \lambda \underbrace{D(\bm{Z}_0,\bm{Z}_1)}_{\mathcal{R}(\phi)},
\end{eqnarray}
where $\bm{Z}_t=\{z_{n,t}\}_{n=1}^{N_t}$ denotes the set of latent representations for the group $t$. 
$D(\bm{Z}_0,\bm{Z}_1)$ denotes the discrepancy between $\bm{Z}_0$ and $\bm{Z}_1$ under a metric $D(\cdot,\cdot)$. 
The first term in~\eqref{eq:classic_cfr} is the loss function measuring the prediction error of the model. 
The second term is the regularizer of the encoder measuring the discrepancy between the latent representations of the two groups, whose significance is controlled by $\lambda>0$. 
This regularizer is often implemented as the special cases of the integral probability metric, e.g., maximum mean discrepancy (MMD)~\cite{shalit2017estimating}, Wasserstein distance~\cite{wang2023optimal}, and so on.
Typically, the order-$2$ Wasserstein distance between $\bm{Z}_0$ and $\bm{Z}_1$ is implemented as
\begin{eqnarray}\label{eq:wd}
\begin{aligned}
    \widehat{W}_2(\bm{Z}_0,\bm{Z}_1)&=\sideset{}{_{\bm{T}\in\Pi(\frac{1}{N_0}\bm{1}_{N_0},\frac{1}{N_1}\bm{1}_{N_1})}}\min\Bigl(\sideset{}{_{m=1}^{N_0}}\sum\sideset{}{_{n=1}^{N_1}}\sum T_{mn}d_Z^2(z_{m,0},z_{n,1})\Bigr)^{\frac{1}{2}}\\
    &=\sideset{}{_{\bm{T}\in\Pi(\frac{1}{N_0}\bm{1}_{N_0},\frac{1}{N_1}\bm{1}_{N_1})}}\min\langle \bm{D}_{Z_{0,1}},\bm{T}\rangle^{\frac{1}{2}},
\end{aligned}
\end{eqnarray}
where the matrix $\bm{D}_{Z_{0,1}}=[d_{Z}^2(z_{m,0},z_{n,1})]\in\mathbb{R}^{N_0\times N_1}$ records the pairwise distances between the latent representations across the two groups.
$d_Z$ is the metric associated with the latent space $\mathcal{Z}$.
$\bm{T}$ is the transport plan matrix, whose feasible domain is $\Pi(\frac{1}{N_0}\bm{1}_{N_0},\frac{1}{N_1}\bm{1}_{N_1})=\{\bm{T}\geq \bm{0}|\bm{T1}_{N_1}=\frac{1}{N_0}\bm{1}_{N_0},\bm{T}^{\top}\bm{1}_{N_0}=\frac{1}{N_1}\bm{1}_{N_1}\}$. 
$\langle\cdot,\cdot\rangle$ denotes inner product. 
The optimal transport plan $\bm{T}^*$ associated with $W_2(\bm{Z}_1,\bm{Z}_2)$ indicates the correspondence between the individuals across the two groups.

As we mentioned before, $\mathcal{R}(\phi)$ aims to suppress the selection bias issue caused by imbalanced $\rho_0$ and $\rho_1$, while existing regularization strategies~\cite{shalit2017estimating,li2022deep} often suffer from the over-enforcing balance issue in their practical implementations. 
In the following content, we will revisit the CFR problem from the viewpoint of information bottleneck, leading to a new learning paradigm mitigating selection bias and over-enforcing balance issues jointly.

\section{Gromov-Wasserstein Information Bottleneck for CFR}\label{sec:method}

\subsection{Motivation}
As shown in~\eqref{eq:pred_ite}, the CFR model is the composition of the encoder $\phi$ and the predictors of the two groups (i.e., $h_0$ and $h_1$). 
From the viewpoint of information bottleneck~\cite{tishby1999information}, we need to achieve a trade-off between the prediction accuracy and the latent representation complexity when learning the model. 
Specifically, for each group, we treat the covariate, its latent representation, and outcome as three random variables, denoted as $\{X_t, Z_t, Y_t\}_{t=0,1}$, and their distributions are $X_t\sim\rho_t$, $Z_t\sim\phi_{\#}\rho_t$, $Y_t\sim p_{Y_t}$, respectively.
The information bottleneck paradigm leads to the following problem:
\begin{eqnarray}\label{eq:ib}
    \sideset{}{_{h_0,h_1,\phi}}\max \sideset{}{_{t=0,1}}\sum I(Z_t, Y_t;h_t,\phi) -\lambda I(Z_t, X_t;\phi).
\end{eqnarray}
Here, $I(A, B;\theta)=\text{KL}(p(A,B)\|p(A)p(B))$ denotes the mutual information between the random variables $A$ and $B$, which can be parameterized by $\theta$.
As shown in~\eqref{eq:ib}, for each group, information bottleneck aims to maximize the mutual information between the latent representation  and the outcome while penalizing the mutual information between the latent representation and the covariate. 
Obviously, the first term aims to pursue the prediction power of the model, and the second term regularizes the complexity of the representation --- the latent representation $Z_t$ should only preserve necessary information from $X_t$ for the prediction task and ignore the useless information.

It is easy to find that the first term in~\eqref{eq:ib} is coincident to the loss function $\mathcal{L}(h_0,h_1,\phi)$ in~\eqref{eq:classic_cfr}.
In particular, the squared loss used in $\mathcal{L}(h_0,h_1,\phi)$ indicates that $p(Y_t|Z_t)=\frac{1}{\sqrt{2\pi}\sigma}\exp(-\frac{1}{2\sigma^2}|h_t(Z_t)-Y_t|^2)$. 
Based on the data $\mathcal{D}_t=\{x_{n,t},y_{n,t}\}_{n=1}^{N_t}$, we have
\begin{eqnarray}\label{eq:first_term}
\begin{aligned}
    &I(Z_t,Y_t;h_t,\phi)=\mathbb{E}_{z,y\sim p(Z_t,Y_t)}[\log p(Y_t=y|Z_t=z) - \log p(Y_t=y)]\\
    &\approx \underbrace{\frac{1}{N_t}\sideset{}{_{n=1}^{N_t}}\sum [\log p(y_{n,t}|z_{n,t}) - \log p_{Y_t}(y_{n,t})]}_{\text{Empirical mutual information:~}\hat{I}_{N_t}(Z_t,Y_t;h_t,\phi))}
    =\frac{-1}{2N_t\sigma^2}\sideset{}{_{n=1}^{N_t}}\sum |h_t\circ\underbrace{\phi(x_{n,t})}_{z_{n,t}}-y_{n,t}|^2 + C,
\end{aligned}
\end{eqnarray}
where $C=-\log\sqrt{2\pi}\sigma-\frac{1}{N}\sum_{n=1}^{N_t}\log p_{Y_t}(y_{n,t})$ is a constant irrelevant to the model. 
Based to~\eqref{eq:first_term}, $\min \mathcal{L}(h_0,h_1,\phi)$ is equivalent to maximizing the empirical mutual information between the latent representations and the outcomes, i.e., 
\begin{eqnarray}
    \min\mathcal{L}(h_0,h_1,\phi)\Leftrightarrow \max \sideset{}{_{t=0,1}}\sum\hat{I}_{N_t}(Z_t,Y_t;h_t,\phi).
\end{eqnarray}

However, the regularizer $\mathcal{R}(\phi)$ in~\eqref{eq:classic_cfr} is inconsistent with the $-\sum_{t=0,1}I(Z_t,X_t;\phi)$ in~\eqref{eq:ib}.
From the viewpoint of information bottleneck, most existing CFR learning methods either ignore or fail to fully consider the penalty of representation complexity.
Therefore, we would like to propose a new regularizer based on the mutual information between the latent representations and the covariates. 

\subsection{Bounding empirical kernelized mutual information}

For the convenience of derivation, we ignore the group's notation ``$t$'' in this subsection without the loss of generality. 
Given $N$ covariates $\{x_n\}_{n=1}^{N}$, we can obtain the corresponding latent representations $\{z_n=\phi(x_n)\}_{n=1}^{N}$ and apply kernel density estimation to obtain their distributions~\cite{chuang2023infoot}, i.e., 
\begin{eqnarray}\label{eq:marginal}
    \rho(X)=\frac{1}{N}\sideset{}{_{n=1}^{N}}\sum\kappa(X, x_{n}),\quad \phi_{\#}\rho(Z)=\frac{1}{N}\sideset{}{_{n=1}^{N}}\sum\kappa(Z, z_{n}),
\end{eqnarray}
where $\kappa(a,b)=\frac{1}{\sqrt{2\pi}\tau}\exp(-\frac{1}{2\tau^2}d^2(a,b))$ is the commonly-used RBF kernel with bandwidth $\tau$ and $d$ is a valid metric.
Accordingly, the joint distribution of $X$ and $Z$ can be estimated based on the observed covariates and their corresponding latent representations, i.e.,
\begin{eqnarray}\label{eq:joint}
    p(X, Z) = \frac{1}{N}\sideset{}{_{n=1}^{N}}\sum \kappa(X,x_{n})\kappa(Z,z_{n}).
\end{eqnarray}
Therefore, we can obtain the empirical kernelized mutual information between $X$ and $Z$ as
\begin{eqnarray}\label{eq:kmi}
    \hat{I}_{\kappa,N}(Z,X;\phi)=\frac{1}{N}\sideset{}{_{n}}\sum\log\frac{p(x_{n},z_{n})}{\rho(x_{n})\phi_{\#}\rho(z_{n})}
    =\frac{1}{N}\sideset{}{_{n}}\sum\log\frac{N \sum_{m}\kappa(x_{n},x_{m})\kappa(z_{n},z_{m})}{\sum_{m}\kappa(x_{n},x_{m})\sum_{m}\kappa(z_{n},z_{m})}.
\end{eqnarray}
Note that applying $\hat{I}_{\kappa,N}(Z,X;\phi)$ directly as the regularizer is questionable in general because the kernel density estimation of $p(X,Z)$ often suffers from the curse of dimensionality.
In practice, the covariates can be high-dimensional variables, so that the observed data $\{x_n,\phi(x_n)\}$ are too sparse to estimate $p(X,Z)$ with high accuracy. 
Specifically, when only $\{x_n,\phi(x_n)\}_{n=1}^{N}$ are available, the coherency of $x_n$ and $z\in\mathcal{Z}\setminus \{\phi({x_n})\}$ and that of $x\in\mathcal{X}\setminus\{x_n\}$ and $\phi(x_n)$ are ignored. 
As a result, the entropy of $(X,Z)$ is underestimated, and $\hat{I}_{\kappa,N}(Z,X;\phi)$ in~\eqref{eq:kmi} becomes unreliable, which prevents the utilization of information bottleneck paradigm.

A potential solution to mitigate this problem is further optimizing the joint distribution $p(X,Z)$ to maximize $\hat{I}_{\kappa,N}(Z,X;\phi)$ and then penalizing the upper bound of $\hat{I}_{\kappa,N}(Z,X;\phi)$ in the information bottleneck paradigm.
Inspired by this idea, we derive an upper bound of $\hat{I}_{\kappa,N}(Z,X;\phi)$, and accordingly, design a surrogate penalty term for~\eqref{eq:ib}, which leads to the proposed Gromov-Wasserstein information bottleneck learning paradigm. 

In particular, our method is based on the following theorem:
\begin{theorem}\label{thm:upperbound}
    Assume that $(\mathcal{X},d_{X})$ and $(\mathcal{Z},d_Z)$ are two bounded spaces, whose diameters are denoted as $\text{Diam}_{\mathcal{X}}$ and $\text{Diam}_{\mathcal{Z}}$, respectively. 
    Given samples $\bm{X}=\{x_{n}\}_{n=1}^{N}$ and corresponding $\bm{Z}=\{z_n\}_{n=1}^N$, 
    for the empirical kernelized mutual information defined in~\eqref{eq:kmi} and using an RBF kernel with bandwidth $\tau$, we have
    \begin{eqnarray}\label{eq:upper}
        \hat{I}_{\kappa,N}(Z,X)\leq \frac{1}{2\tau^2}\Bigl(\frac{1}{N^2}\|\bm{D}_{X}-\bm{D}_{Z}\|_F^2 - \widehat{GW}_2^2(\bm{X},\bm{Z})\Bigr)+C_{\kappa,N},
    \end{eqnarray}
    where $\bm{D}_{X}=[d_X(x_{m},x_{n})]\in\mathbb{R}^{N\times N}$ and $\bm{D}_{Z}=[d_Z(z_{m},z_{n})]\in\mathbb{R}^{N\times N}$ are distance matrices, 
    \begin{eqnarray}\label{eq:gwd_sample}
        \widehat{GW}_2(\bm{X},\bm{Z})=\sideset{}{_{\bm{T}\in \Pi(\frac{1}{N}\bm{1}_{N},\frac{1}{N}\bm{1}_{N})}}\min\langle\bm{C}_{XZ}-2\bm{D}_{X}\bm{T}\bm{D}_{Z}^{\top},~\bm{T}\rangle^{\frac{1}{2}}.
    \end{eqnarray}
    is the order-$2$ Gromov-Wasserstein discrepancy\footnote{The formulation in~\eqref{eq:gwd_sample} is based on the Proposition 1 in~\cite{peyre2016gromov}, which is applicable when applying the RBF kernel in the derivation of~\eqref{eq:upper}.} between $\bm{X}$ and $\bm{Z}$, where $\bm{C}_{XZ}=\frac{1}{N}((\bm{D}_{X}\odot\bm{D}_{X})\bm{1}_{N\times N}+\bm{1}_{N\times N}(\bm{D}_{Z}\odot\bm{D}_{Z})^{\top})$.
    $C_{\kappa,N}=\frac{N^4(2\pi\tau^2-\alpha)^2}{8(N^4-1)\alpha^2}$, where $\alpha=\exp(-\frac{\text{Diam}^2_{\mathcal{X}}+\text{Diam}^2_{\mathcal{Z}}}{2\tau^2})$.
\end{theorem}
\textit{Sketched Proof.} For the empirical kernelized mutual information $\hat{I}_{\kappa,N}(Z,X)$, we first apply Jensen's inequality to the term $-\frac{1}{N}\sum_{n}\log \rho(x_{n})\phi_{\#}\rho(z_{n})$, leading to the first term of the upper bound.
Then, based on the work in~\cite{chuang2023infoot}, we consider the optimal joint distribution of $\{x_m, z_n\}_{m,n=1}^{N}$ that maximizes the empirical kernelized mutual information and relax the term $\frac{1}{N_t}\sum_{n}\log p(x_{n},z_{n})$ to $C_{\kappa}-\frac{1}{2\tau^2}\widehat{GW}_2^2(\bm{X},\bm{Z})$, where $C_{\kappa,N}$ is the upper bound of the Jensen gap for the logarithmic function in a closed set~\cite{costarelli2015sharp}. 
A detailed proof is in Appendix~\ref{appendix:thm}.

\textbf{Remark.} 
Real-world data such as healthcare records or online user profiles rarely contain infinite or unbounded covariates. 
When applying bounded activation layers in the encoder $\phi$, we can obtain bounded latent representations.
Therefore, it is reasonable for us to assume $(\mathcal{X},d_{X})$ and $(\mathcal{Z},d_Z)$ to be bounded in Theorem~\ref{thm:upperbound}.

\subsection{Connections to Monge gap and Monge map approximation}
The upper bound in~\eqref{eq:upper} corresponds the Gromovization of the Monge gap proposed in~\cite{uscidda2023monge}.
\begin{definition}[Gromovized Monge gap]\label{def-gromov-monge-gap}
    Let $(\mathcal{X},d_{X})$ and $(\mathcal{Z},d_{Z})$ be two compact metric spaces, respectively.
    $\rho$ is a probability measure defined on $(\mathcal{X},d_{X})$, and $\phi:\mathcal{X}\mapsto\mathcal{Z}$ is a mapping function. 
    The order-$p$ Gromovized Monge gap associated with $\phi$ and $\rho$ is defined as 
    \begin{eqnarray}\label{eq:gmg}
    \begin{aligned}
        MG_{GW_p}(\rho,\phi)=\mathbb{E}_{x,x'\sim\rho\times\rho}[|d_X(x,x')-d_Z(\phi(x),\phi(x'))|^p] - GW_p^p(\rho,\phi_{\#}\rho),
    \end{aligned}
    \end{eqnarray}
    where the first term measures the transport cost of pushing $\rho$ to $\phi_{\#}\rho$ through $\phi$, and the second term $GW_p(\rho,\phi_{\#}\rho)$ is the order-$p$ GW distance between $\rho$ and $\phi_{\#}\rho$, i.e.,
    \begin{eqnarray*}\label{eq:gwd}
    \begin{aligned}
    GW_p(\rho,\phi_{\#}\rho)=\sideset{}{_{\pi\in\Pi(\rho,\phi_{\#}\rho)}}\inf\Bigl(\int_{\mathcal{X}^2\times\mathcal{Z}^2}|d_X(x,x')-d_Z(z,z')|^p\mathrm{d}\pi(x,z)\mathrm{d}\pi(x',z')\Bigr)^{\frac{1}{p}}.
    \end{aligned}
    \end{eqnarray*}
\end{definition}
The Gromovized Monge gap replaces the Wasserstein distance used in the original Monge gap~\cite{uscidda2023monge} with the GW distance, making this concept applicable for the Monge map across different spaces. 
It measures the discrepancy between the transport cost of pushing $\rho$ to $\phi_{\#}\rho$ and the GW distance~\cite{memoli2011gromov} between $\rho$ and $\phi_{\#}\rho$. 
According to the above definition, we can equivalently rewrite~\eqref{eq:upper} as
\begin{eqnarray}\label{eq:upper2}
    \hat{I}_{\kappa,N}(Z,X;\phi) = \mathcal{O}(MG_{GW_2}(\hat{\rho}_N,\phi)),
\end{eqnarray}
where $\hat{\rho}_{N}$ denotes the empirical distribution of the observed covariates $\{x_n\}_{n=1}^N$.
When $N\rightarrow\infty$, we have $\widehat{GW}_2(\bm{X},\bm{Z})\rightarrow GW_2(\rho,\phi_{\#}\rho)$, $\frac{1}{N^2}\|\bm{D}_{X}-\bm{D}_{Z}\|_F^2\rightarrow \mathbb{E}_{x,x'\sim \rho\times\rho}[|d_X(x,x')-d_Z(\phi(x),\phi(x'))|^2]$, and $C_{\kappa,N}\rightarrow \frac{(2\pi\tau-\alpha)^2}{8\alpha^2}$. 
Therefore, $\hat{I}_{\kappa,N}(Z,X;\phi)=\mathcal{O}(MG_{GW_2}(\rho,\phi))$, which means that the upper bound we derived is consistent with the increase of sample size.

Similar to~\cite{uscidda2023monge}, we can prove that 
\begin{theorem}\label{thm:gmg}
    $MG_{GW_p}(\rho,\phi)\geq 0$ and $MG_{GW_p}(\rho,\phi)=0$ if and only if $\phi$ is the Monge map between $\rho$ and $\phi_{\#}\rho$. 
\end{theorem}
It means that the upper bound of $\hat{I}_{\kappa,N}(Z,X;\phi)$ is minimized when $\phi:\mathcal{X}\mapsto\mathcal{Z}$ is the Monge map pushing $\hat{\rho}_{N}$ to $\phi_{\#}\hat{\rho}_{N}$.

\subsection{Incorporating with OT-based balancing regularization}

Based on Theorem~\ref{thm:upperbound} and~\eqref{eq:upper2}, we can take $\sum_{t=0,1}MG_{GW_2}(\hat{\rho}_{t,N_t},\phi)$ as the penalty term when learning the CFR model. 
Moreover, by further relaxing the upper bound, we can easily introduce the classic balancing regularizer into the GWIB paradigm.
Denote $\frac{\|\bm{D}_{X_t}-\bm{D}_{Z_t}\|_F^2}{N_t^2}$ as $R_t^2$.
For each group $t$, we can relax $MG_{GW_2}(\hat{\rho}_{t,N_t},\phi)$ as follows: 
\begin{eqnarray}\label{eq:upper3}
\begin{aligned}
    &MG_{GW_2}(\hat{\rho}_{t,N_t},\phi)=(R_t-\widehat{GW}_2(\bm{X}_{t},\bm{Z}_{t}))(R_t+\widehat{GW}_2(\bm{X}_{t},\bm{Z}_{t}))\\
    &\leq (R_t-\widehat{GW}_2(\bm{X}_{t},\bm{Z}_{1-t})+\widehat{GW}_2(\bm{Z}_{t},\bm{Z}_{1-t}))(R_t+\widehat{GW}_2(\bm{X}_{t},\bm{Z}_{1-t})+\widehat{GW}_2(\bm{Z}_{t},\bm{Z}_{1-t}))\\
    &=(R_t + \widehat{GW}_2(\bm{Z}_{t},\bm{Z}_{1-t}))^2-\widehat{GW}_2^2(\bm{X}_{t},\bm{Z}_{1-t})\\
    &\leq \Bigl(R_t + \frac{1}{\beta}\widehat{FGW}_{2,\beta}(\bm{Z}_{t},\bm{Z}_{1-t})\Bigr)^2-\widehat{GW}_2^2(\bm{X}_{t},\bm{Z}_{1-t}),
\end{aligned}
\end{eqnarray}
where $\widehat{FGW}_{2,\beta}(\bm{Z}_{t},\bm{Z}_{1-t})$ is the order-$2$ fused Gromov-Wasserstein (FGW) distance~\cite{titouan2019optimal}, i.e., 
\begin{eqnarray}\label{eq:fgwd}
    \sideset{}{_{\bm{T}\in\Pi(\frac{1}{N_t}\bm{1}_{N_t},\frac{1}{N_{1-t}}\bm{1}_{N_{1-t}})}}\min((1-\beta)\underbrace{\langle\bm{D}_{Z_{t,1-t}},\bm{T}\rangle}_{\text{Wass. term in~\eqref{eq:wd}}}+\beta\underbrace{\langle\bm{C}_{Z_tZ_{1-t}}-2\bm{D}_{Z_t}\bm{T}\bm{D}_{Z_{1-t}}^{\top},\bm{T}\rangle}_{\text{GW term}})^{\frac{1}{2}},
\end{eqnarray}
where $\beta\in [0,1]$ achieves the trade-off between the Wasserstein term and the GW term.

In~\eqref{eq:upper3}, we apply the triangle inequality to obtain the first inequality because GW distance is a valid metric~\cite{memoli2011gromov}.
The second inequality in~\eqref{eq:upper3} is based on the fact that $\beta \widehat{GW}_2(\bm{Z}_{t},\bm{Z}_{1-t})\leq \widehat{FGW}_{2,\beta}(\bm{Z}_{t},\bm{Z}_{1-t})$.
Similar to~\eqref{eq:wd}, the FGW distance also captures the discrepancy between $\bm{Z}_t$ and $\bm{Z}_{1-t}$, which works as an OT-based balancing regularization.

\textbf{Remark.} Compared with combining $\widehat{FGW}_{2,\beta}(\bm{Z}_{t},\bm{Z}_{1-t})$ (or $\widehat{W}_2(\bm{Z}_{t},\bm{Z}_{1-t})$) with $\sum_{t=0,1}MG_{GW_2}(\hat{\rho}_{t,N_t},\phi)$, our relaxation method has an advantage on inferring consistent cross-group individual correspondence. 
In particular, the $\widehat{GW}_2(\bm{X}_t,\bm{Z}_t)$ in the $MG_{GW_2}(\hat{\rho}_{t,N_t},\phi)$ computes the optimal transport plan for the individuals within the group $t$, while the optimal transport plan associated with $\widehat{FGW}_{2,\beta}(\bm{Z}_{t},\bm{Z}_{1-t})$ (or $\widehat{W}_2(\bm{Z}_{t},\bm{Z}_{1-t})$) is defined for the individuals across the two groups. 
Therefore, the na\"{i}ve regularization $\widehat{FGW}_2(\bm{Z}_{t},\bm{Z}_{1-t};\beta)+\sum_{t=0,1}MG_{GW_2}(\hat{\rho}_{t,N_t},\phi)$ needs to optimize three irrelevant optimal transport plans.
On the other hand, applying our relaxation strategy in~\eqref{eq:upper3}, we can implement the proposed regularizer as
\begin{eqnarray}\label{eq:reg}
    \mathcal{R}(\phi,\bm{T}_{0,1}^*,\bm{T}_{0}^*,\bm{T}_1^*)=\sideset{}{_{t=0,1}}\sum\Bigl(R_t + \frac{1}{\beta}\widehat{FGW}_{2,\beta}(\bm{Z}_{0},\bm{Z}_{1};\bm{T}_{0,1}^*)\Bigr)^2-\widehat{GW}_2^2(\bm{X}_{t},\bm{Z}_{1-t};\bm{T}_{t}^*),
\end{eqnarray}
where $\widehat{FGW}_{2,\beta}(\bm{Z}_{0},\bm{Z}_{1};\bm{T}_{0,1}^*)$ denotes the FGW term associated with the OT plan $\bm{T}_{0,1}^*\in\Pi(\frac{1}{N_0}\bm{1}_{N_0},\frac{1}{N_{1}}\bm{1}_{N_{1}})$, and similarly,  $\widehat{GW}_2^2(\bm{X}_{t},\bm{Z}_{1-t};\bm{T}_{t}^*)$ denotes the GW term associated with the OT plan $\bm{T}_t^*\in\Pi(\frac{1}{N_t}\bm{1}_{N_t},\frac{1}{N_{1-t}}\bm{1}_{N_{1-t}})$ for $t=0,1$. 
All the three OT plans indicate cross-group individual correspondence, which should be consistent with each other.

Focusing on CFR, we summarize the functionality of each term in~\eqref{eq:reg}: $i)$ $\widehat{FGW}_{2,\beta}(\bm{Z}_{0},\bm{Z}_{1};\bm{T}_{0,1}^*)$ captures the discrepancy between the latent representations of different groups, so penalizing this term mitigates the selection bias issue. $ii)$ $R_t=\frac{1}{N_t}\|\bm{D}_{X_t}-\bm{D}_{Z_t}\|_F$ captures the discrepancy between the covariate pairs and the corresponding latent representation pairs, so penalizing $R_t$ helps avoid generating potential trivial solutions caused by over-enforced balancing.
$iii)$ $\widehat{GW}_2^2(\bm{X}_{t},\bm{Z}_{1-t};\bm{T}_{t}^*)$ helps infer cross-group individual correspondence, and penalizing the gap between this term and $(R_t+\widehat{FGW}_{2,\beta}(\bm{Z}_{0},\bm{Z}_{1};\bm{T}_{0,1}^*))^2$ helps control the representation complexity of the encoder $\phi$.

Applying the regularizer in~\eqref{eq:reg}, our GWIB paradigm learns the CFR model by solving the following bi-level optimization problem:
\begin{eqnarray}\label{eq:gwib}
\begin{aligned}
    \sideset{}{_{h_0,h_1,\phi}}\min&~\mathcal{L}(h_0,h_1,\phi) + \lambda \mathcal{R}(\phi,\bm{T}_{0,1}^*,\bm{T}_{0}^*,\bm{T}_1^*),\\
    s.t.~&\bm{T}_t^*\in\arg\sideset{}{_{\bm{T}\in\Pi(\frac{1}{N_t}\bm{1}_{N_t},\frac{1}{N_{1-t}}\bm{1}_{N_{1-t}})}}\min \langle \bm{C}_{X_tZ_{1-t}}-2\bm{D}_{X_t}\bm{T}\bm{D}_{Z_{1-t}}^{\top},\bm{T}\rangle,~t=0,1,\\
    &\bm{T}_{0,1}^*\in\arg\sideset{}{_{\bm{T}\in\Pi(\frac{1}{N_0}\bm{1}_{N_0},\frac{1}{N_{1}}\bm{1}_{N_{1}})}}\min \langle (1-\beta)\bm{D}_{Z_{0,1}}+\beta\bm{C}_{Z_0Z_1}-2\beta\bm{D}_{Z_0}\bm{T}\bm{D}_{Z_1}^{\top},\bm{T}\rangle,\\
    &\bm{T}_{0}^*=(\bm{T}_{1}^*)^{\top}=\bm{T}_{0,1}^*,
\end{aligned}
\end{eqnarray}
where the first three constraints correspond to the computation of the GW and FGW terms, while the last equality constraint guarantees the consistency of the cross-group individual correspondence.
With the help of this constraint, we can simplify~\eqref{eq:gwib} and obtain the final learning problem as follows:
\begin{eqnarray}\label{eq:gwib2}
\begin{aligned}
    &\sideset{}{_{h_0,h_1,\phi}}\min~\mathcal{L}(h_0,h_1,\phi) + \lambda \mathcal{R}(\phi,\bm{T}^*,\bm{T}^*,(\bm{T}^*)^{\top}),\\
    &s.t.~\bm{T}^*\in\arg\sideset{}{_{\bm{T}\in\Pi(\frac{1}{N_0}\bm{1}_{N_0},\frac{1}{N_{1}}\bm{1}_{N_{1}})}}\min \langle \bm{C}_{X_0Z_1}-2\bm{D}_{X_0}\bm{T}\bm{D}_{Z_1}^{\top}+\bm{C}_{X_1Z_0}^{\top}-2\bm{D}_{Z_0}^{\top}\bm{T}\bm{D}_{X_1}\\
    &\hspace{5.5cm}+(1-\beta)\bm{D}_{Z_{0,1}}+\beta\bm{C}_{Z_0Z_1}-2\beta\bm{D}_{Z_0}\bm{T}\bm{D}_{Z_1}^{\top},~\bm{T}\rangle,
\end{aligned}
\end{eqnarray}
where the proposed regularizer in~\eqref{eq:reg} takes $\{\bm{T}^*,\bm{T}^*,(\bm{T}^*)^{\top}\}$ as its input OT plans.

\subsection{Optimization Algorithm}\label{sec-learn}
Leveraging alternating optimization, we design an effective algorithm to solve~\eqref{eq:gwib} iteratively.
In each iteration, the proposed algorithm involves the following two steps.

\textbf{Update $\bm{T}$.} Given the current encoder $\phi$, we first solve the lower-level problem in~\eqref{eq:gwib2} by the well-known conditional gradient (CG) algorithm~\cite{titouan2019optimal}. 
Denote the objective function of the lower-level problem as $\mathcal{L}(\bm{T})$. 
We first compute the gradient $\nabla_{\bm{T}}\mathcal{L}$ as
\begin{eqnarray*}  
\bm{C}_{X_0Z_1}-2\bm{D}_{X_0}\bm{T}\bm{D}_{Z_1}^{\top}+\bm{C}_{X_1Z_0}^{\top}-2\bm{D}_{Z_0}^{\top}\bm{T}\bm{D}_{X_1}+(1-\beta)\bm{D}_{Z_{0,1}}+\beta\bm{C}_{Z_0Z_1}-2\beta\bm{D}_{Z_0}\bm{T}\bm{D}_{Z_1}^{\top},
\end{eqnarray*}
and then update $\bm{T}$ by projected gradient descent, i.e., $\bm{T}\leftarrow \text{Proj}_{\Pi(\frac{1}{N_0}\bm{1}_{N_0},\frac{1}{N_{1}}\bm{1}_{N_{1}})}(\bm{T}-\bm{S}\odot\nabla_{\bm{T}}\mathcal{L})$, where the elementwise step size $\bm{S}$ is determined by line search~\cite{dessein2018regularized}.
Repeating the above steps till $\mathcal{L}(\bm{T})$ converges, we obtain the optimal $\bm{T}^*$ given $\phi$.
Although this problem is nonsmooth and nonconvex, the CG algorithm makes it converge to a local stationary point~\cite{lacoste2016convergence}.

\textbf{Update model parameters.} Given the current $\bm{T}^*$, we plug it into the proposed regularizer and update $\{\phi,h_0,h_1\}$ by solving $\min_{\phi,h_0,h_1}\mathcal{L}(\phi,h_0,h_1)+\lambda\mathcal{R}(\phi,\bm{T}^*,\bm{T}^*,(\bm{T}^*)^{\top})$. 
The classic mini-batch stochastic gradient descent (SGD) is applied.

Repeating the above two steps till the objective function converges, we obtain the proposed model.
We refer the readers to Appendix~\ref{appendix:opt} for more optimization details.

\section{Related Work}
\textbf{ITE estimation and counterfactual regression.} 
Recently, representation-based methods~\cite{johansson2016learning} are well-studied for ITE estimation, which map individuals' covariates into a latent space and predict counterfactual outcomes based on the latent representations. 
As a typical representation-based method, counterfactual regression (CFR)~\cite{shalit2017estimating} minimizes the prediction error of observed outcomes and regularizes the discrepancy (e.g., Wasserstein distance) between the latent distributions jointly. 
The regularization in the CFR framework can be implemented in various ways, including applying different distance metrics (e.g., CFR-MMD~\cite{shalit2017estimating}, DTANet~\cite{li2022deep} and ESCFR~\cite{wang2023optimal}) and re-sampling (or re-weighting) underrepresented subpopulations (e.g., StableCFR~\cite{wu2023stable}).

Among the above methods, some optimal transport-based methods achieved encouraging performance in various causal inference tasks~\cite{charpentier2023optimal,li2021causal,dunipace2021optimal,torous2021optimal}.
Focusing on ITE estimation, DTANet~\cite{li2022deep} applies entropic optimal transport (EOT) distance as the regularizer, and ESCFR~\cite{wang2023optimal} leverages unbalanced optimal transport (UOT) to address the misalignment issues caused by outliers and unobserved confounders.

\textbf{Information bottleneck and its applications.}
Information bottleneck (IB) principle~\cite{tishby1999information} helps balance representation complexity and prediction accuracy when learning model, which has been widely used in many learning tasks and practical applications~\cite{harremoes2007information,ozair2019wasserstein}. 
Some attempts have been made to connect IB to optimal transport (OT) theory. 
The work in~\cite{chen2023information} reformulates the IB paradigm as an entropic OT problem. 
The InfoOT framework in~\cite{chuang2023infoot} maximizes kernelized mutual information by learning an OT plan associated with the joint distribution of variables.
In the field of causal inference, the work in~\cite{parbhoo2020information} utilizes IB to compress covariate information and then transfer it to the test samples with missing features.
The work in~\cite{kim2019reliable,lu2022causal} propose causal IB paradigms for causal effect prediction.

Different from the above work, our GWIB provides a new OT-based technique route to implement the information bottleneck paradigm, leading to a new CFR method for ITE prediction.
Our method $i)$ bounds kernelized empirical mutual information between covariate and corresponding latent representation with Gromovized Monge gap, and
$ii)$ incorporates the FGW distance as a balancing penalty.
The encoder approximates Monge maps, which suppresses selection bias and over-enforcing balance jointly, and the OT plan indicates the consistent cross-group individual correspondence.

\section{Experiments}

\subsection{Experimental Setup}
\label{sec:exp-set}
\textbf{Datasets}.
Following prior work~\cite{wang2023optimal}, we conduct experiments on two semi-synthetic datasets, IHDP and ACIC.  
IHDP~\cite{hill2011bayesian} comprises 747 samples with 25-dimensional covariates from a real-world randomized experiment, while ACIC~\cite{dorie2019automated} includes data for 4802 patients with 58 covariates.

\vspace{0.2cm}
\noindent
\textbf{Methods}.
We consider three groups of methods as baselines: 
$i)$ meta-learners (\textbf{S-Learner} and \textbf{T-Learner}~\cite{kunzel2019metalearners}),
$ii)$ matching-based methods (\textbf{$k$-NN}~\cite{crump2008nonparametric} and \textbf{PSM}~\cite{caliendo2008some}),
and $iii)$ representation-based methods (\textbf{TARNet}, \textbf{CFR-Wass}, \textbf{CFR-MMD}~\cite{shalit2017estimating}, \textbf{DragonNet}~\cite{shi2019adapting}, \textbf{SITE}~\cite{yao2018representation}, \textbf{BWCFR}~\cite{assaad2021counterfactual} and \textbf{ESCFR}~\cite{wang2023optimal}).
We compare our proposed \textbf{GWIB} method with all these baselines~\footnote{The source code and data are available at https://github.com/peteryang1031/Causal-GWIB.}.

\vspace{0.2cm}
\noindent
\textbf{Evaluations}. 
We employ the absolute error in average treatment effect $\epsilon_{ATE}$, and the precision in estimation of heterogeneous effects $\epsilon_{PEHE}$ as metrics to assess the precision of treatment effect estimation.
Additionally, we conduct both in-sample and out-sample experiments to evaluate the generalizability of our method. 
More results and implementation details can be found in Appendix~\ref{appendix-setup}.

\begin{table*}[t]
\caption{Overall comparison results on ACIC and IHDP datasets.}
\label{exp-main}
\LARGE
{
\resizebox{\textwidth}{!}{
    \begin{tabular}{c|cccc|cccc}		
        \hline 
        \hline
        Datasets&\multicolumn{4}{c|}{ACIC}&\multicolumn{4}{c}{IHDP} \\ \midrule
        Test Types&\multicolumn{2}{c}{In-sample}&\multicolumn{2}{c|}{Out-sample}&\multicolumn{2}{c}{In-sample}&\multicolumn{2}{c}{Out-sample} \\ \midrule
        Methods & $\epsilon_{ATE}$ & $\epsilon_{PEHE}$  & $\epsilon_{ATE}$ & $\epsilon_{PEHE}$& $\epsilon_{ATE}$ & $\epsilon_{PEHE}$& $\epsilon_{ATE}$ & $\epsilon_{PEHE}$ \\ \midrule
        S-Learner & 0.8120$_{\pm \text{0.2220}}$ & 2.2790$_{\pm \text{0.0340}}$ & 1.0137$_{\pm \text{0.2844}}$ & 2.2997$_{\pm \text{0.0493}}$ & 0.4032$_{\pm \text{0.5535}}$ & 1.2304$_{\pm \text{0.4826}}$ & 0.3629$_{\pm \text{0.3130}}$ & 0.9359$_{\pm \text{0.2040}}$ \\ 
        T-Learner & 0.9086$_{\pm \text{0.0779}}$ & 1.7873$_{\pm \text{0.0131}}$ & 2.0749$_{\pm \text{0.1887}}$ & 2.6557$_{\pm \text{0.1270}}$ & 0.3038$_{\pm \text{0.1356}}$ & 1.3700$_{\pm \text{0.0412}}$ & 0.3305$_{\pm \text{0.3468}}$ & 1.4256$_{\pm \text{0.1844}}$ \\ \midrule
        $k$-NN & 0.9005$_{\pm \text{0.0007}}$ & 4.9456$_{\pm \text{0.0004}}$ & 0.7432$_{\pm \text{0.0085}}$ & 5.0947$_{\pm \text{0.0011}}$ & 1.1671$_{\pm \text{0.0224}}$ & 3.3651$_{\pm \text{0.0283}}$ & 1.3891$_{\pm \text{0.0592}}$ & 2.9556$_{\pm \text{0.0762}}$ \\ 
        PSM & 0.8712$_{\pm \text{0.0000}}$ & 4.7940$_{\pm \text{0.0003}}$ & 0.6153$_{\pm \text{0.0001}}$ & 4.9540$_{\pm \text{0.0006}}$ & 1.0764$_{\pm \text{0.0000}}$ & 3.2496$_{\pm \text{0.0000}}$ & 0.7785$_{\pm \text{0.0245}}$ & 2.1364$_{\pm \text{0.0000}}$ \\ \midrule
        TARNet & 0.7775$_{\pm \text{0.2063}}$ & 1.7215$_{\pm \text{0.1005}}$ & 1.6519$_{\pm \text{0.9499}}$ & 2.4702$_{\pm \text{0.9361}}$ & 0.1311$_{\pm \text{0.0510}}$ & 0.8711$_{\pm \text{0.0387}}$ & 0.0827$_{\pm \text{0.0539}}$ & 0.8944$_{\pm \text{0.0686}}$ \\ 
        DragonNet & 0.5160$_{\pm \text{0.2251}}$ & 1.6614$_{\pm \text{0.1269}}$ & 0.7845$_{\pm \text{0.6219}}$ & 2.1005$_{\pm \text{0.6729}}$ & 0.1355$_{\pm \text{0.0054}}$ & 0.8752$_{\pm \text{0.0020}}$ & 0.1078$_{\pm \text{0.0052}}$ & 0.8989$_{\pm \text{0.0017}}$ \\ 
        CFR-Wass & 0.4213$_{\pm \text{0.0125}}$ & 1.5291$_{\pm \text{0.0079}}$ & 0.7470$_{\pm \text{0.3123}}$ & 1.7889$_{\pm \text{0.1082}}$ & 0.1057$_{\pm \text{0.0265}}$ & 0.8700$_{\pm \text{0.0678}}$ & 0.0597$_{\pm \text{0.0346}}$ & 0.9142$_{\pm \text{0.0529}}$ \\ 
        CFR-MMD & 0.4156$_{\pm \text{0.2296}}$ & 1.3874$_{\pm \text{0.0673}}$ & 0.9183$_{\pm \text{0.5629}}$ & 1.7446$_{\pm \text{0.3479}}$ & 0.1073$_{\pm \text{0.0100}}$ & 0.8650$_{\pm \text{0.0843}}$ & 0.0565$_{\pm \text{0.0490}}$ & 0.8859$_{\pm \text{0.1034}}$ \\ 
        ESCFR & 0.4155$_{\pm \text{0.0292}}$ & 1.4091$_{\pm \text{0.0048}}$ & 1.0831$_{\pm \text{0.1972}}$ & 1.8048$_{\pm \text{0.0811}}$ & 0.1015$_{\pm \text{0.0008}}$ & 0.8315$_{\pm \text{0.0066}}$ & 0.1135$_{\pm \text{0.0026}}$ & 0.8854$_{\pm \text{0.0096}}$ \\ 
        SITE & 0.8716$_{\pm \text{0.3252}}$ & 2.4311$_{\pm \text{0.1879}}$ & 0.7173$_{\pm \text{0.2787}}$ & 2.3388$_{\pm \text{0.0285}}$ & 0.1132$_{\pm \text{0.1044}}$ & 0.9766$_{\pm \text{0.0656}}$ & 0.1004$_{\pm \text{0.0854}}$ & 0.8398$_{\pm \text{0.1229}}$ \\ 
        BWCFR & 0.6391$_{\pm \text{0.2422}}$ & 1.5775$_{\pm \text{0.0664}}$ & 1.2175$_{\pm \text{0.2302}}$ & 1.9100$_{\pm \text{0.1161}}$ & 0.1391$_{\pm \text{0.0436}}$ & 0.8592$_{\pm \text{0.0911}}$ & 0.0517$_{\pm \text{0.0224}}$ & 0.8639$_{\pm \text{0.0843}}$ \\ \midrule
        GWIB (Ours) & \textbf{0.3252}$_{\pm \text{0.0498}}$ & \textbf{1.2542}$_{\pm \text{0.0045}}$ & \textbf{0.5163}$_{\pm \text{0.2786}}$ & \textbf{1.4527}$_{\pm \text{0.0389}}$ & \textbf{0.0506}$_{\pm \text{0.0006}}$ & \textbf{0.5085}$_{\pm \text{0.0003}}$ & \textbf{0.0503}$_{\pm \text{0.0003}}$ & \textbf{0.6988}$_{\pm \text{0.0002}}$ \\
        \hline \hline
    \end{tabular}
    }
}
\end{table*}

\begin{figure}[t]
\setlength{\abovecaptionskip}{0.2cm}
\setlength{\fboxrule}{0.pt}
\setlength{\fboxsep}{0.pt}
\centering
    \subfigure{
        \includegraphics[width=0.33\textwidth]{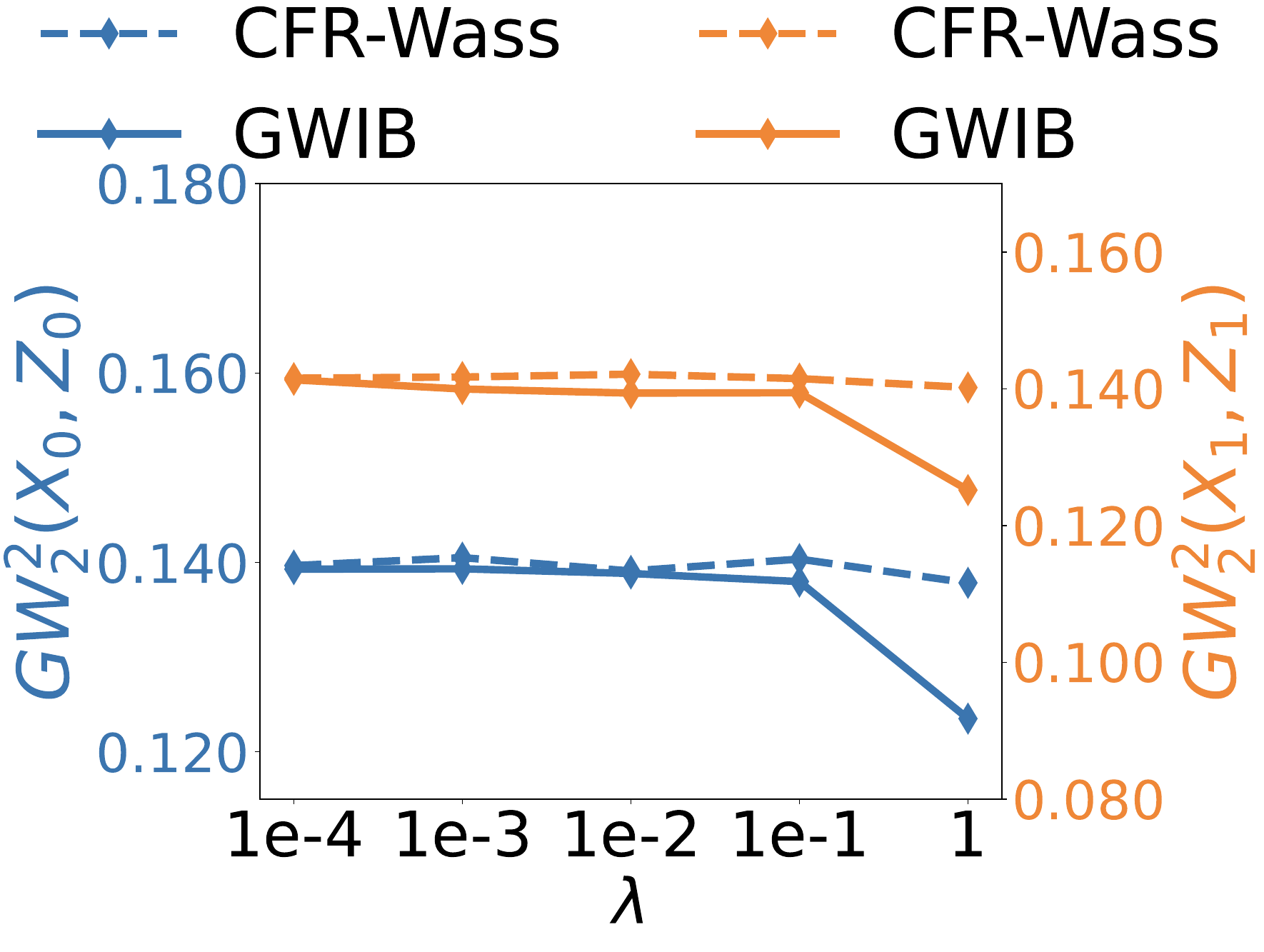}	
    }
    \subfigure{
        \includegraphics[width=0.33\textwidth]{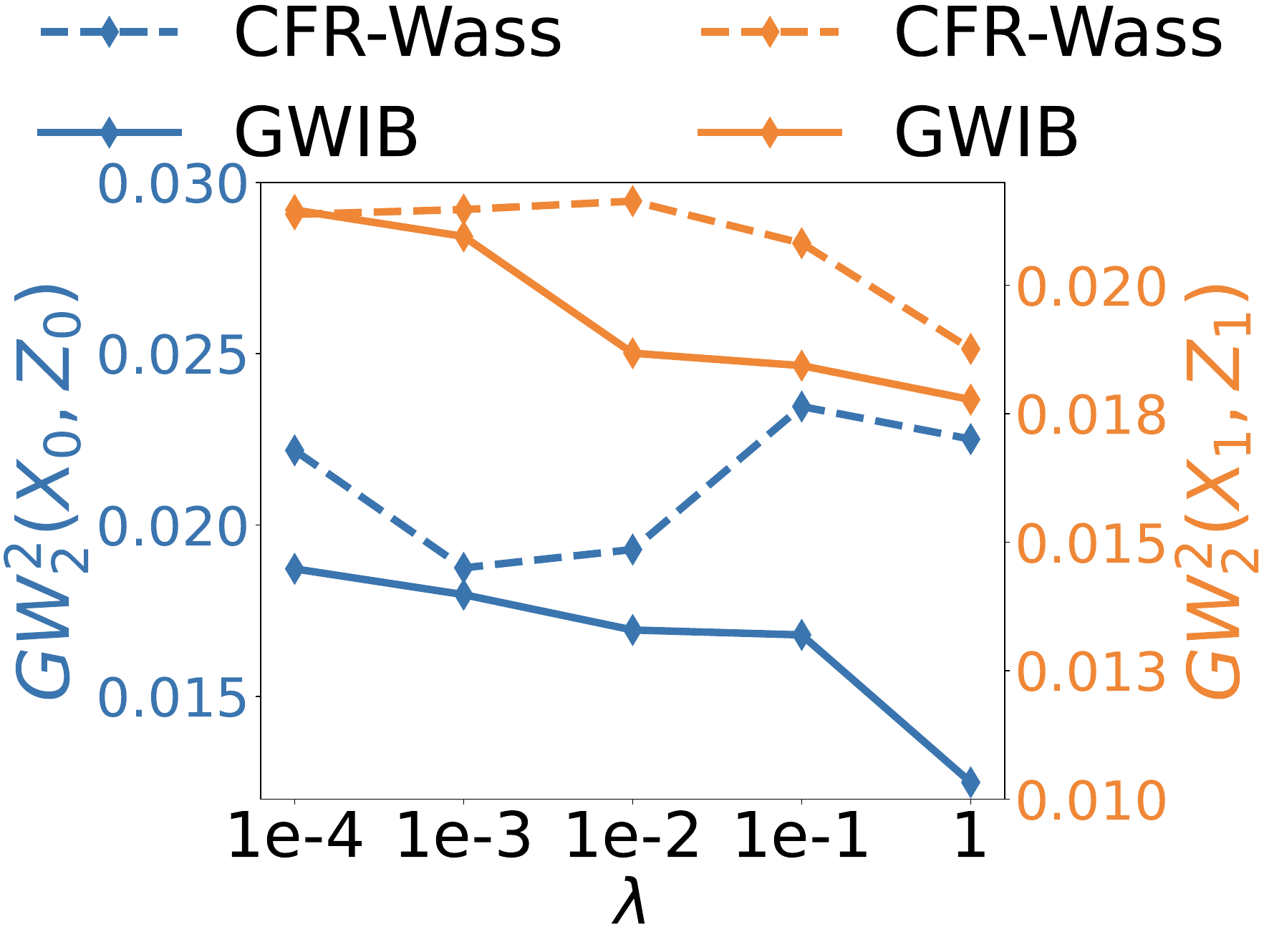}
    }\\
\subfigure{
    \includegraphics[width=0.33\textwidth]{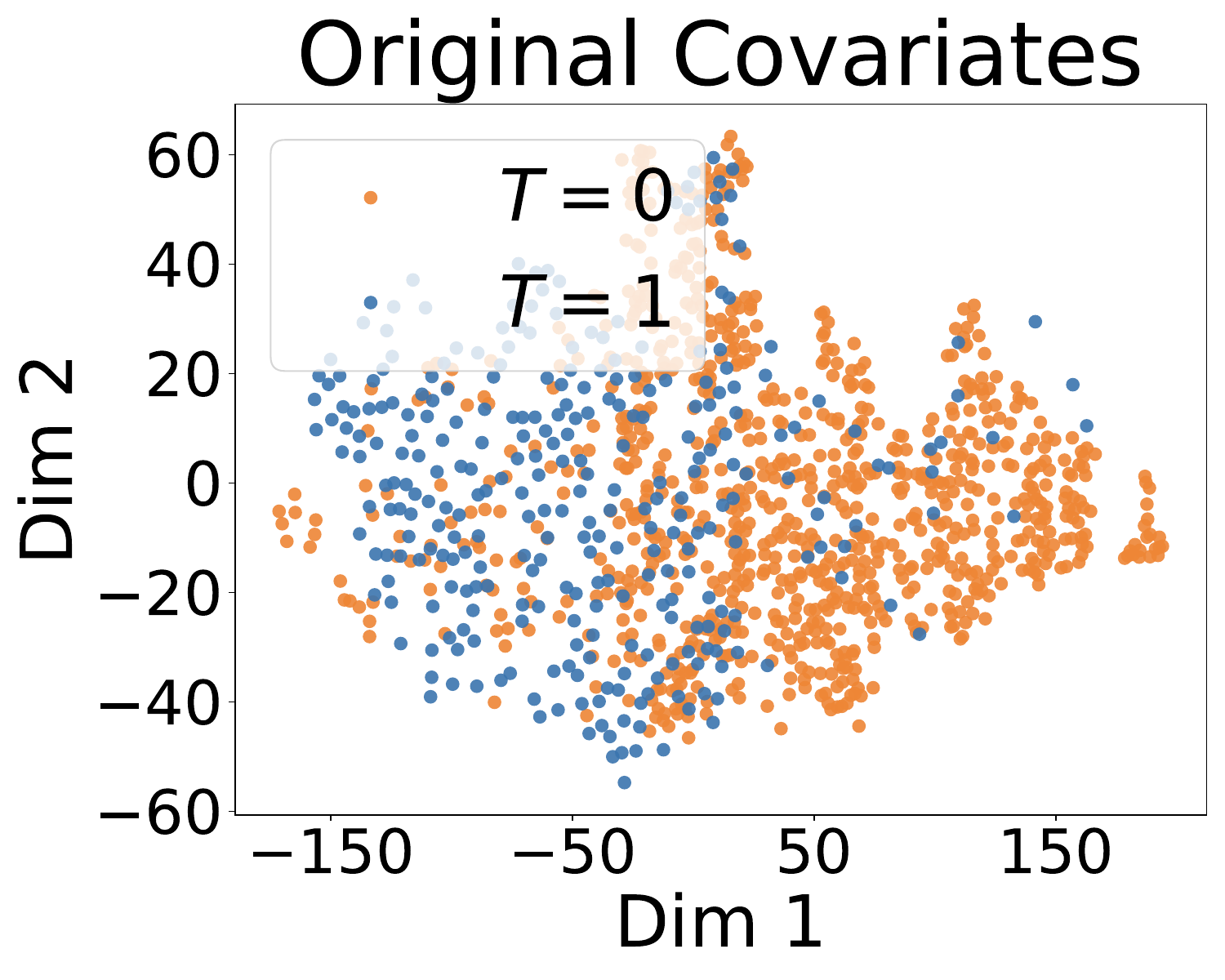}	
}
\hspace{-0.25cm}
\subfigure{
    \includegraphics[width=0.3\textwidth]{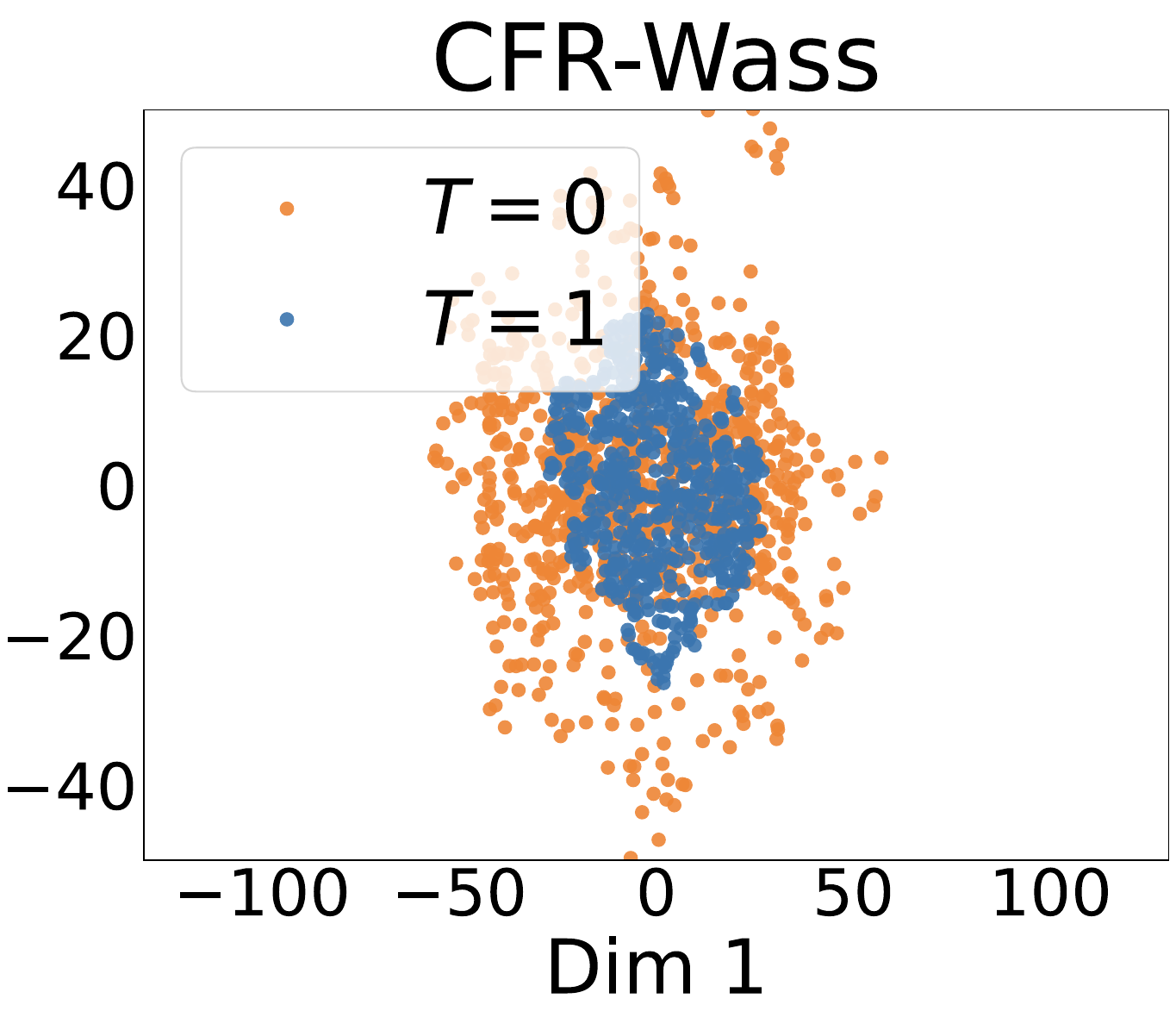}
}
\hspace{-0.25cm}
\subfigure{
    \includegraphics[width=0.3\textwidth]{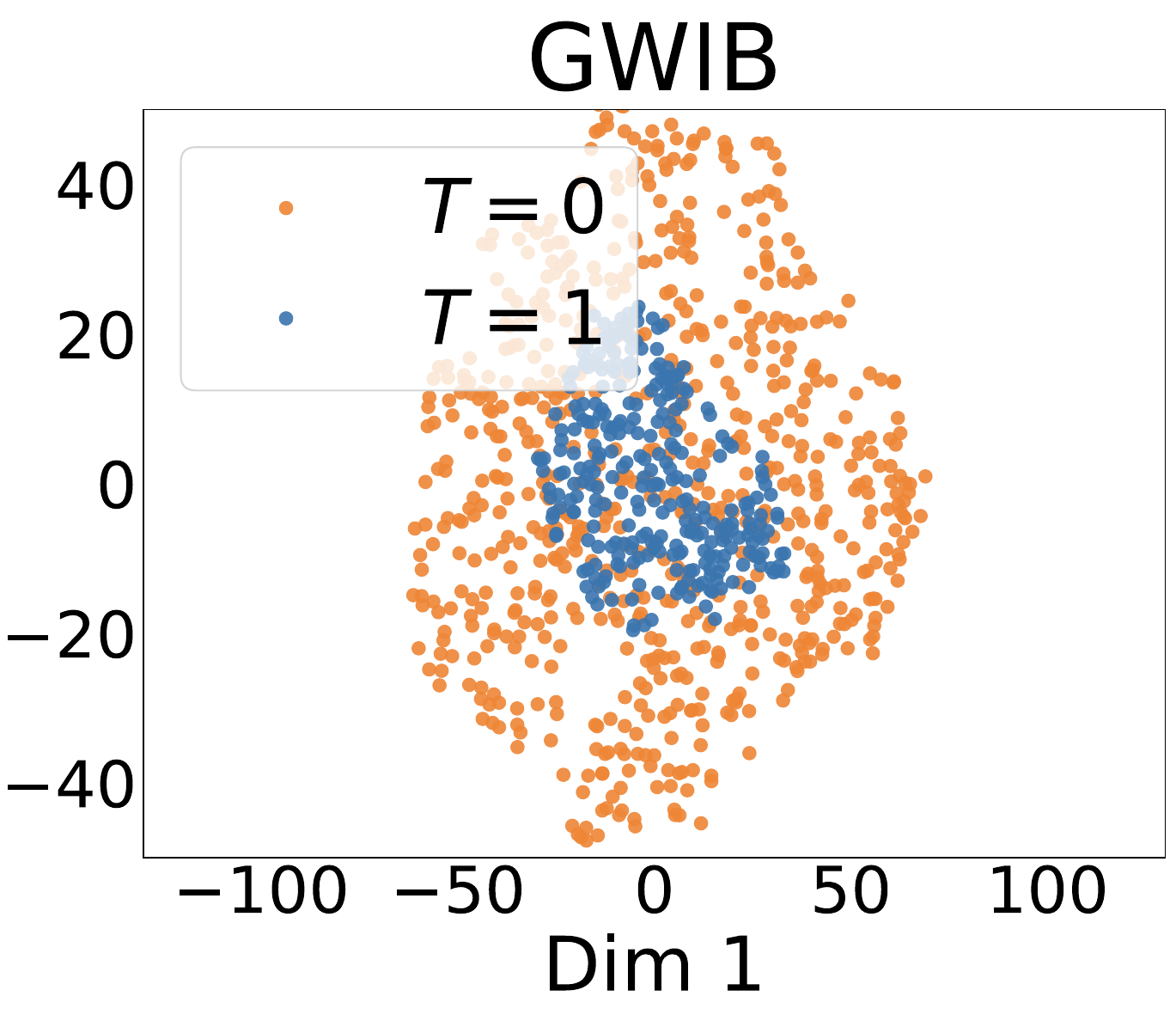}	
}
    \caption{(Upper) Illustrations of the GW distances between the covariate distributions and the latent counterparts on ACIC and IHDP datasets. (Lower) The t-SNE plots of the original covariates of ACIC and the latent representations achieved by CFR-Wass and GWIB.}\label{fig:tsne_gw}
\end{figure}

\begin{table*}[t]
    \caption{Ablation studies on ACIC and IHDP datasets. The best results are highlighted by bold font.}
    \label{exp-ablation}
    \LARGE{
    \resizebox{\textwidth}{!}{
        \begin{tabular}{c|cccc|cccc}
            \hline 
            \hline
            Datasets&\multicolumn{4}{c|}{ACIC}&\multicolumn{4}{c}{IHDP} \\ \midrule
            Test Types&\multicolumn{2}{c}{In-sample}&\multicolumn{2}{c|}{Out-sample}&\multicolumn{2}{c}{In-sample}&\multicolumn{2}{c}{Out-sample} \\ \midrule
            Methods & $\epsilon_{ATE}$ & $\epsilon_{PEHE}$  & $\epsilon_{ATE}$ & $\epsilon_{PEHE}$& $\epsilon_{ATE}$ & $\epsilon_{PEHE}$& $\epsilon_{ATE}$ & $\epsilon_{PEHE}$ \\ \midrule
            GWIB-FGW & 0.3752$_{\pm \text{0.0562}}$ & 1.2852$_{\pm \text{0.0066}}$ & 1.0038$_{\pm \text{0.2803}}$ & 1.6396$_{\pm \text{0.1064}}$ & 0.1530$_{\pm \text{0.0110}}$ & 0.6605$_{\pm \text{0.0746}}$ & 0.1580$_{\pm \text{0.0067}}$ & 0.8974$_{\pm \text{0.0067}}$ \\ 
            GWIB-RT & 0.3806$_{\pm \text{0.1747}}$ & 1.4222$_{\pm \text{0.0217}}$ & 1.1034$_{\pm \text{0.2940}}$ & 1.7818$_{\pm \text{0.1218}}$ & 0.1079$_{\pm \text{0.0018}}$ & 0.5681$_{\pm \text{0.0108}}$ & 0.1824$_{\pm \text{0.0071}}$ & 0.7889$_{\pm \text{0.0249}}$ \\ 
            GWIB-GW & 0.4580$_{\pm \text{0.0311}}$ & 1.2882$_{\pm \text{0.0036}}$ & 1.0764$_{\pm \text{0.3035}}$ & 1.6758$_{\pm \text{0.1020}}$ & 0.1304$_{\pm \text{0.0032}}$ & 0.5561$_{\pm \text{0.0058}}$ & 0.1684$_{\pm \text{0.0025}}$ & 0.8083$_{\pm \text{0.0187}}$ \\ 
            GWIB-Gap & 0.3906$_{\pm \text{0.0406}}$ & 1.2922$_{\pm \text{0.0038}}$ & 1.2205$_{\pm \text{0.2043}}$ & 1.7770$_{\pm \text{0.0756}}$ & 0.1248$_{\pm \text{0.0016}}$ & 0.5156$_{\pm \text{0.0001}}$ & 0.1503$_{\pm \text{0.0062}}$ & 0.7174$_{\pm \text{0.0002}}$ \\ 
            GWIB-Opt & 0.5060$_{\pm \text{0.0041}}$ & 1.2991$_{\pm \text{0.0033}}$ & 1.3675$_{\pm \text{0.0586}}$ & 1.8378$_{\pm \text{0.0349}}$ & 0.1326$_{\pm \text{0.0043}}$ & 0.5234$_{\pm \text{0.0002}}$ & 0.1809$_{\pm \text{0.0067}}$ & 0.7170$_{\pm \text{0.0006}}$ \\ \midrule
            GWIB (Ours) & \textbf{0.3252}$_{\pm \text{0.0498}}$ & \textbf{1.2542}$_{\pm \text{0.0045}}$ & \textbf{0.5163}$_{\pm \text{0.2786}}$ & \textbf{1.4527}$_{\pm \text{0.0389}}$ & \textbf{0.0506}$_{\pm \text{0.0006}}$ & \textbf{0.5085}$_{\pm \text{0.0003}}$ & \textbf{0.0503}$_{\pm \text{0.0003}}$ & \textbf{0.6988}$_{\pm \text{0.0002}}$ \\
            \hline \hline
        \end{tabular}
    }
    }
\end{table*}

\subsection{Comparisons on ITE Estimation}
\label{sec:exp-ite}
The main results on ACIC and IHDP datasets are reported in Table~\ref{exp-main}, demonstrating the superiority of our GWIB method clearly.
In particular, our method outperforms the baselines consistently in all prediction tasks, where we have the following observations:
$i)$ Meta-learners, despite being parameterized by neural networks to capture non-linearity, exhibit inferior performance in estimating treatment effects since they do not address the treatment selection bias.
$ii)$ Matching-based methods achieve suboptimal performance on $\epsilon_{PEHE}$ due to the absence of regularization on covariate distributions. 
Instead, they only search for similar individuals in the opposite treatment group as counterfactuals, making it challenging to achieve unbiased results at the distribution level.
$iii)$ Compared to existing representation-based methods, our GWIB achieves the best results in all metrics. 
This is because these baselines fail to penalize the mutual information between covariates and their latent representations, and thus generate trivial solutions with inconsistent cross-group correspondence.

We attribute the superior performance of GWIB to its ability to mitigate selection bias and over-enforcing balance jointly at the distribution level --- making the transport map approximate Monge map prevents the collapse of covariate distributions. 
To verify the above claim, we apply our regularization with different strengths (i.e., setting $\lambda\in [1e^{-4}, 1]$) to learn the transport map $\phi$ and use the order-$2$ Gromov-Wasserstein distance between each covariate distribution and its latent counterpart (i.e., $\widehat{GW}_2^2(\bm{X}_0, \bm{Z}_0)$ and $\widehat{GW}_2^2(\bm{X}_1, \bm{Z}_1)$) to measure the information loss caused by the transport map.
As shown in the upper of Figure~\ref{fig:tsne_gw}, GWIB consistently exhibits lower GW distances than CFR-Wass in both datasets. 
Additionally, for the ACIC dataset, we visualize the t-SNE plots of the original convariates and the latent representations obtained by CFR-Wass and GWIB in the lower of Figure~\ref{fig:tsne_gw}.
The results clearly show that both CFR-Wass and GWIB can successfully balance the distributions of latent representations, mitigating treatment selection bias between the control group and treatment group. 
However, GWIB method achieves balanced latent distributions with large supports, effectively mitigating the over-enforcing balance, while CFR-Wass displays collapsed latent distributions as it does not penalize the loss of information induced by the transport map.

\subsection{Ablation Studies}
Our Gromov-Wasserstein based information bottleneck paradigm is implemented with several components, as shown in~\eqref{eq:reg}. 
To analyze the contributions of these components quantitatively, we conduct several ablation studies by tailoring the following five variants of GWIB: 
$i)$ \underline{GWIB-FGW} removes the FGW distance in~\eqref{eq:reg}; 
$ii)$ \underline{GWIB-RT} removes the $R^t$ terms in~\eqref{eq:reg};
$iii)$ \underline{GWIB-GW} removes the GW distances in~\eqref{eq:reg};
$iv)$ \underline{GWIB-Gap} optimizes the original Monge gap in the first line in~\eqref{eq:upper3};
$v)$ \underline{GWIB-Opt} computes the OT plans of the GW and FGW terms independently during optimization.
All of these variants are fed into the same two-head regression frameworks for a fair comparison.

The results presented in Table~\ref{exp-ablation} demonstrate the superior performance of our GWIB method compared to its five variants across all experiments. 
Specifically, the comparisons for our GWIB and the first four variants show the necessity of applying the complete Gromov-Wasserstein information bottleneck regularization.
The comparison to GWIB-Opt shows the significance of learning consistent cross-group individual correspondence by a single OT plan.

\section{Conclusion and Future Work}
In this paper, we revisit CFR through the lens of a novel Gromov-Wasserstein information bottleneck paradigm, leading to a strong ITE estimation method that suppresses selection bias and over-enforcing balance issues jointly.
The proposed method connects the upper bound of kernelized empirical mutual information with a Gromovized Monge gap, which optimizes the CFR encoder to approximate the Monge maps from covariate distributions to latent ones.
Experiments in various ITE estimation tasks demonstrate the effectiveness of our method. 
Our work pioneers the theoretical connection between information bottleneck and counterfactual regression methods, providing a new optimal transport-based technical route.
In the future, we plan to apply the proposed regularization method to other causal inference applications, e.g., networked ITE estimation~\cite{guo2020learning}, dynamic ITE estimation~\cite{ma2021deconfounding} and sequential ITE estimation~\cite{bica2020estimating}.

\bibliographystyle{plainnat}
\bibliography{bib}
\newpage
\appendix

\section{Proofs of Theorems}\label{appendix:thm}
\subsection{Proof of Theorem~\ref{thm:upperbound}}\label{appendix:thm}

\textbf{A generalized KDE of the joint probability $p(X,Z)$.}
Recall the joint probability $p(X,Z)$ defined in~\eqref{eq:joint}. 
We can find that it is actually a special case of the following generalized formulation proposed in~\cite{chuang2023infoot}.
\begin{eqnarray}\label{eq:infoot}
    p(X,Z)=\sideset{}{_{m,n=1}^{N}}\sum T_{mn}\kappa(X,x_m)\kappa(Z,z_n),~\text{with}~\bm{T}=[T_{mn}]\in\Pi\Bigl(\frac{1}{N}\bm{1}_N,\frac{1}{N}\bm{1}_N\Bigr),
\end{eqnarray}
where $\bm{T}$ represents the joint distribution of the sample pairs $(x_m,z_n)$.
Obviously, the joint distribution in~\eqref{eq:joint} corresponds to the case with $\bm{T}=\frac{1}{N}\bm{I}_N$, and we have
\begin{eqnarray}\label{eq:general_kde}
    p(X,Z)\leq \sideset{}{_{\bm{T}\in\Pi(\frac{1}{N}\bm{1}_N,\frac{1}{N}\bm{1}_N)}}\max \sideset{}{_{m,n=1}^{N}}\sum T_{mn}\kappa(X,x_m)\kappa(Z,z_n).
\end{eqnarray}

\textbf{The Jensen gap for the logarithmic function defined on a closed domain.}
Suppose that $x\in (0,\infty)$ and obeys to a distribution $p$.
For $\log x$, Jensen's inequality tells us that  $\log\mathbb{E}_{x\sim p}[x]\geq \mathbb{E}_{x\sim p}[\log x]$, and the $\log\mathbb{E}_{x\sim p}[x]-\mathbb{E}_{x\sim p}[\log x]$ is called ``Jensen gap''. 

The work in~\cite{costarelli2015sharp} studies the sharpness of the Jensen's inequality, deriving the upper bound of Jensen gap for various functions, including $\log x$:
\begin{lemma}[The upper bound of Jensen gap for logarithmic function in~\cite{costarelli2015sharp}]\label{lemma:gap}
Suppose that $\log x$ is defined on a closed set $[a,\infty)$, where $a>0$, and $x\sim p$, whose mean is $\mu$ and variance is $\sigma^2$. 
We have
\begin{eqnarray}\label{eq:gap}
    \log\mathbb{E}_{x\sim p}[x]-\mathbb{E}_{x\sim p}[\log x]\leq\frac{1}{2a^2}\min_{c\geq a}[\mathbb{E}_{x\sim p}[(x-c)^2]+(\mu-c)^2]=\frac{\sigma^2}{2a^2}. 
\end{eqnarray}
\end{lemma}
Obviously, this lemma is held when $\log x$ is defined in $[a, b]$, $b<\infty$. 

Given $\{x_m,z_n\}_{m,n=1}^{N}\in\mathcal{X}\times\mathcal{Z}$, where $(\mathcal{X},d_X)$ and $(\mathcal{Z},d_Z)$ are bounded metric spaces with $\text{diam}_{\mathcal{X}}=\sup_{x,x'}d_X(x,x')$ and $\text{diam}_{\mathcal{Z}}=\sup_{z,z'}d_Z(z,z')$. 
The range of the function $f(x,x',z,z')=\kappa(x,x')\kappa(z,z')$ is a closed interval $[\frac{1}{2\pi\tau^2}\alpha, 1]$, for $m,n=1,...,N$, where $\alpha=\exp(-\frac{1}{2\tau^2}(\text{diam}_{\mathcal{X}}^2+\text{diam}_{\mathcal{Z}}^2))$.

Considering the generalized formulation of KDE shown in~\eqref{eq:infoot}, we can take $T_{m,n}T_{k,l}$ as the probability of $f(x_m,x_k,z_n,z_l)$. 
Then, we can estimate the mean and variance of $f(x,x',z,z')$ based on $\{x_m,z_n\}_{m,n=1}^N$ (i.e., $N^4$ $(x,x',z,z')$ samples in total), using the classic moment estimation method, i.e.,
\begin{eqnarray}
\begin{aligned}
    &\hat{\mu}=\sum_{m,n,k,l=1}^N T_{m,n}T_{k,l} f(x_m,x_k,z_n,z_l),\\
    &\hat{\sigma}^2=\frac{N^4}{N^4-1}\sum_{m,n,k,l=1}^N T_{m,n}T_{k,l} (f(x_m,x_k,z_n,z_l) - \hat{\mu})^2\leq \frac{N^4(1-\frac{\alpha}{2\pi\tau^2})^2}{4(N^4-1)},
\end{aligned}
\end{eqnarray}
where $\frac{N^4}{N^4-1}$ is the coefficient ensuring unbiased variance estimation.
The upper bound of $\hat{\sigma}_{m,n}^2$ is achieved when a half of $\{f(x_m,x_k,z_n,z_l)\}_{m,n,k,l=1}^N$ equal to $1$ while the remaining samples equal to $\frac{1}{2\pi\tau^2}\exp(-\frac{1}{2\tau^2}(\text{diam}_{\mathcal{X}}^2+\text{diam}_{\mathcal{Z}}^2))$. 

Based on the upper bound of the variance and Lemma~\ref{lemma:gap}, we have
\begin{eqnarray}\label{eq:c_kappa}
\begin{aligned}
    &\log\Bigl(\sum_{m,n,k,l}T_{m,n}T_{k,l}\kappa(x_m,x_k)\kappa(z_n,z_l)\Bigr)\\
    &\leq \sum_{m,n,k,l}T_{m,n}T_{k,l}\log(\kappa(x_m,x_k)\kappa(z_n,z_l)) + \underbrace{\frac{N^4(2\pi\tau^2-\alpha)^2}{8(N^4-1)\alpha^2}}_{C_{\kappa,N}},
\end{aligned}    
\end{eqnarray}
where the last term is the proposed Jensen gap in this work.

According to the above information, we have
\begin{eqnarray*}
\begin{aligned}
    &\hat{I}_{\kappa,N}(Z,X)\\
    =&\frac{1}{N}\sum_{n}\Bigl(\log\Bigl(\frac{1}{N}\sum_{m}\kappa(x_n,x_m)\kappa(z_n,z_m)\Bigr) -\log\Bigl(\frac{1}{N^2}\sum_{m}\kappa(x_n,x_m)\sum_{m}\kappa(z_n,z_m)\Bigr) \Bigr)\\
    \leq &\log\Bigl(\frac{1}{N^2}\sum_{m,n}\kappa(x_m,x_n)\kappa(z_m,z_n)\Bigr)-\frac{1}{N^2}\sum_{m,n}\Bigl(\log\kappa(x_m,x_n)+\log\kappa(z_m,z_n)\Bigr)\\
    =&\log\Bigl(\frac{1}{N^2}\sum_{m,n}\kappa(x_m,x_n)\kappa(z_m,z_n)\Bigr)+\frac{\sum_{m,n}(d_X^2(x_m,x_n)+d_Z^2(z_m,z_n))}{2N^2\tau^2}+\log2\pi\tau^2\\
    =&\log\Bigl(\frac{1}{N^2}\sum_{m,n}\kappa(x_m,x_n)\kappa(z_m,z_n)\Bigr)+\frac{\sum_{m,n}d_X(x_m,x_n)d_Z(z_m,z_n)}{N^2\tau^2}+\frac{\sum_{m,n}(d_X(x_m,x_n)-d_Z(z_m,z_n))^2}{2N^2\tau^2}+\log2\pi\tau^2\\
    \leq&\max_{\bm{T}\in\Pi(\frac{1}{N}\bm{1}_N,\frac{1}{N}\bm{1}_N)}\log\Bigl(\sum_{m,n,k,l=1}^{N}T_{m,n}T_{k,l}\kappa(x_m,x_k)\kappa(z_n,z_l)\Bigr)+\frac{\sum_{m,n,k,l}T_{m,n}T_{k,l}d_X(x_m,x_k)d_Z(z_n,z_l)}{\tau^2}\Bigr)\\
    &+\frac{1}{2N^2\tau^2}\|\bm{D}_X-\bm{D}_Z\|_F^2+\log2\pi\tau^2\\
    \leq &\max_{\bm{T}\in\Pi(\frac{1}{N}\bm{1}_N,\frac{1}{N}\bm{1}_N)}\sum_{m,n,k,l=1}^{N}T_{m,n}T_{k,l}\Bigl(\log\Bigl(\kappa(x_m,x_k)\kappa(z_n,z_l)\Bigr)+\frac{d_X(x_m,x_k)d_Z(z_n,z_l)}{\tau^2}\Bigr)+\frac{\|\bm{D}_X-\bm{D}_Z\|_F^2}{2N^2\tau^2}+\log2\pi\tau^2\\
    &+C_{\kappa,N}\\
    = & \max_{\bm{T}\in\Pi(\frac{1}{N}\bm{1}_N,\frac{1}{N}\bm{1}_N)}\frac{-1}{2\tau^2}\sum_{m,n,k,l}T_{m,n}T_{k,l}(d_X(x_m,x_k)-d_Z(z_n,z_l))^2+\frac{\|\bm{D}_X-\bm{D}_Z\|_F^2}{2N^2\tau^2}+C_{\kappa,N}\\
    =&\frac{1}{2\tau^2}\Bigl(\frac{\|\bm{D}_X-\bm{D}_Z\|_F^2}{N^2}- \min_{\bm{T}\in\Pi(\frac{1}{N}\bm{1}_N,\frac{1}{N}\bm{1}_N)}\sum_{m,n,k,l}T_{m,n}T_{k,l}(d_X(x_m,x_k)-d_Z(z_n,z_l))^2 \Bigr) + C_{\kappa,N}\\
    =&\frac{1}{2\tau^2}\Bigl(\frac{1}{N^2}\|\bm{D}_X-\bm{D}_Z\|_F^2-GW_2^2(\bm{X},\bm{Z}) \Bigr)+C_{\kappa,N},
\end{aligned}    
\end{eqnarray*}
which completes the proof.
Here, the first inequality is derived based on Jensen's inequality. 
The second inequality is based on the upper bound of the generalized KDE shown in~\eqref{eq:general_kde}, which generalizes $\bm{T}=\frac{1}{N}\bm{I}_N$ to $\bm{T}\in\Pi(\frac{1}{N}\bm{1}_N,\frac{1}{N}\bm{1}_N)$.
The third inequality is based on the upper bound of Jensen gap shown in~\eqref{eq:c_kappa}.

\subsection{Proof of Theorem~\ref{thm:gmg}}
Specifically, the GM distance is defined as follows:
\begin{definition}[Gromov-Monge distance]\label{def-gromov-monge}
Let $\mu$ and $\nu$ be two probability measures defined on two compact metric spaces $(\mathcal{X},d_{X})$ and $(\mathcal{Z},d_{Z})$, respectively. 
The order-$p$ Gromov-Monge distance between $\mu$ and $\nu$ is
\begin{eqnarray}\label{eq:gmd}
\begin{aligned}
GM_p(\mu,\nu):&=\inf_{\psi_{\#}\mu=\nu}\Bigl(\int_{\mathcal{X}^2}r^p(x,x',\psi(x),\psi(x'))\mathrm{d}\mu(x)\mathrm{d}\mu(x')\Bigr)^{\frac{1}{p}}\\
&=\inf_{\psi_{\#}\mu=\nu}\mathbb{E}_{x,x'\sim \mu\times\mu}^{\frac{1}{p}}[r^p(x,x',\psi(x),\psi(x'))],\\
\end{aligned}
\end{eqnarray}
where the relational distance $r(x,x',\psi(x),\psi(x'))=|d_{X}(x,x')-d_Z(\psi(x),\psi(x'))|$.
The infimum is over all functions $\psi:\mathcal{X}\mapsto\mathcal{Z}$ that push $\mu$ to $\nu$, i.e., $\psi_{\#}\mu=\nu$, and the function $\psi^*$ achieving the infimum is called Monge map.
\end{definition}

Similar to the Monge gap in~\cite{uscidda2023monge}, the positivity of Gromov-Monge gap is also based on the fact~\cite{memoli2022distance} that
\begin{eqnarray}
    GM_p(\mu,\nu)\geq GW_{p}(\mu, \nu),~\forall ~\mu\in\mathbb{P}(\mathcal{X}),~\nu\in\mathbb{P}(\mathcal{Z}).
\end{eqnarray}
When the equality holds if and only if the optimal transport plan $\pi^*$ associated with $GW_{p}(\mu, \nu)$ is induced by the Monge map $\psi^*$ associated with $GM_p(\mu,\nu)$, i.e., $\pi=(id,\psi)_{\#}\mu$.

\section{Optimization}
\begin{algorithm}[tb]
\caption{Bi-level Optimization of GWIB}
\label{alg:main}
\begin{algorithmic}[1]
	\Require  $\{x_m,y_m(x_m,0)\}_{m=1}^{N_0}$,  $\{x_n,y_n(x_n,1)\}_{n=1}^{N_1}$, $\hat{\rho}_0 = \{x_m\}^{N_0}_{m=0}, \hat{\rho}_1=\{x_n\}_{n=1}^{N_1}$, $\beta, \lambda$, learning rate $\eta$, number of iterations $K$, number of epochs $M$.
	\For{$m=1, \dots, M$}
	\Statex \textsc{\underline{Update OT Plan $\bm{T}$:}}
    \State Compute cost matrices $\bm{D}_{X_0}=[\|x_m-x_n\|_2^2]\in\mathbb{R}^{N_0\times N_0}$, $\bm{D}_{X_1}=[\|x_m-x_n\|_2^2]\in\mathbb{R}^{N_1\times N_1}$.
	\State Compute latent cost matrices $\bm{D}_{Z_0}=[\|\phi(x_m)-\phi(x_n)\|_2^2] \in\mathbb{R}^{N_0\times N_0} $, $\bm{D}_{Z_1}=[\|\phi(x_m)-\phi(x_n)\|_2^2] \in\mathbb{R}^{N_1\times N_1}$, 
    $\bm{D}_{Z_{0,1}}=[\|\phi(x_m)-\phi(x_n)\|_2^2] \in\mathbb{R}^{N_0\times N_1}$.
	\State $\bm{T}^0 \leftarrow \hat{\rho}_0 \hat{\rho}_1^\top$.
	\For{$k=1, \dots, K$}
	\State $\nabla_{\bm{T}}\mathcal{L} \leftarrow$ Compute gradient w.r.t. $\bm{T}^{(k-1)}$. 
	\State $\widetilde{\bm{T}}^{(k)} \leftarrow$ Solve OT problem with gradient $\nabla_{\bm{T}}\mathcal{L}$. 
	\State $\tau^{(k)} \leftarrow$ Line-Search($\beta, \widetilde{\bm{T}}^{(k)},\bm{T}^{(k-1)}, \bm{D}_{X_0},\bm{D}_{X_1}, \bm{D}_{Z_0}, \bm{D}_{Z_1}, \bm{D}_{Z_{0,1}}$) in Alg.~\ref{alg:ls}.
	\State $\bm{T}^{(k)} \leftarrow (1 - \tau^{(k)}) \bm{T}^{(k-1)} + \tau^{(k)}\widetilde{\bm{T}}^{(k)}$. 
	\EndFor
	\State $\widehat{\mathcal{R}}(\phi,\bm{T}^*) \leftarrow$ Compute loss of~\eqref{eq:reg} w.r.t. $\bm{T}^{(K)}$. 
	\Statex \textsc{\underline{Update parameters $\{\phi, h_0, h_1\}$: }} 
	\State $Y_0 \leftarrow \{y_m(x_m, 0)\}_{m=1}^{N_0}$, $Y_1 \leftarrow \{y_n(x_n, 1)\}_{n=1}^{N_1}$.
	\State $\hat{Y}_0 \leftarrow h_0 \circ \phi(\hat{\rho}_0), \hat{Y}_1 \leftarrow h_1 \circ \phi(\hat{\rho}_1)$. 
	\State $\mathcal{L}(\phi, h_0, h_1) \leftarrow$ Compute squared loss of~\eqref{eq:classic_cfr} w.r.t. $\hat{Y}_0, \hat{Y}_1, Y_0, Y_1$.
	\State $\phi \leftarrow \phi  - \eta \frac{\partial (\mathcal{L}(\phi, h_0, h_1) + \lambda \mathcal{R}(\phi, \bm{T}^*))}{\partial \phi}$.
	\State $h_0 \leftarrow h_0  - \eta \frac{\partial (\mathcal{L}(\phi, h_0, h_1)}{\partial h_0}$.
	\State $h_1 \leftarrow h_1  - \eta \frac{\partial (\mathcal{L}(\phi, h_0, h_1)}{\partial h_1}$.
	\EndFor \\
	\Return $\{\phi, h_0, h_1\}$.
\end{algorithmic}
\end{algorithm}
\begin{algorithm}[t]
\caption{Line-Search}
\label{alg:ls}
\begin{algorithmic}[1]
	\Require  $\beta, \widetilde{\bm{T}}^{(k)},\bm{T}^{(k-1)}, \bm{D}_{X_0},\bm{D}_{X_1}, \bm{D}_{Z_0}, \bm{D}_{Z_1}, \bm{D}_{Z_{0,1}}$.
	\Ensure $\tau^{(k)}$.
        \State $\bm{C}_{X_0Z_1}=\frac{1}{N_0}(\bm{D}_{X_0}\odot\bm{D}_{X_0})\bm{1}_{N_0\times N_1}+\frac{1}{N_1}\bm{1}_{N_0\times N_1}(\bm{D}_{Z_1}\odot\bm{D}_{Z_1})$.
        \State $\bm{C}_{X_1Z_0}=\frac{1}{N_1}(\bm{D}_{X_1}\odot\bm{D}_{X_1})\bm{1}_{N_1\times N_0}+\frac{1}{N_0}\bm{1}_{N_1\times N_0}(\bm{D}_{Z_0}\odot\bm{D}_{Z_0})$.
        \State $\bm{C}_{Z_0Z_1}=\frac{1}{N_0}(\bm{D}_{Z_0}\odot\bm{D}_{Z_0})\bm{1}_{N_0\times N_1}+\frac{1}{N_1}\bm{1}_{N_0\times N_1}(\bm{D}_{Z_1}\odot\bm{D}_{Z_1})$.
	\State $a = -2\langle \beta \bm{D}_{Z_0}\widetilde{\bm{T}}^{(k)}\bm{D}_{Z_1} + \bm{D}_{Z_0}\widetilde{\bm{T}}^{(k)}\bm{D}_{X_1} + \bm{D}_{X_0}\widetilde{\bm{T}}^{(k)}\bm{D}_{Z_1}, \widetilde{\bm{T}}^{(k)}\rangle$.
	\State $b = \langle (1-\beta) \bm{D}_{Z_{0,1}} + \beta \bm{C}_{Z_0Z_1} + \bm{C}_{X_0Z_1} + \bm{C}_{X_1Z_0}^\top, \widetilde{\bm{T}}^{(k)} \rangle-2(\langle \beta \bm{D}_{Z_0}\widetilde{\bm{T}}^{(k)}\bm{D}_{Z_1} + \bm{D}_{Z_0}\widetilde{\bm{T}}^{(k)}\bm{D}_{X_1} + \bm{D}_{X_0}\widetilde{\bm{T}}^{(k)}\bm{D}_{Z_1}, \bm{T}^{(k-1)} \rangle)
	-2(\langle \beta \bm{D}_{Z_0}\bm{T}^{(k-1)}\bm{D}_{Z_1} + \bm{D}_{Z_0}\bm{T}^{(k-1)}\bm{D}_{X_1} + \bm{D}_{X_0}\bm{T}^{(k-1)}\bm{D}_{Z_1}, \widetilde{\bm{T}}^{(k)} \rangle).$
	\If{$a > 0$}
	\State $\tau^{(k)} \leftarrow \min(1, \max (0, \frac{-b}{2a}))$.
	\Else
	\State $\tau^{(k)} \leftarrow 1 \  \text{if} \  a + b < 0 \  \text{else} \  \tau^{(k)} \leftarrow 0$.
	\EndIf
	\State \Return $\tau^{(k)}$.
\end{algorithmic}
\end{algorithm}
\label{appendix:opt}
\begin{figure}[t]
	\setlength{\abovecaptionskip}{0.2cm}
	\setlength{\fboxrule}{0.pt}
	\setlength{\fboxsep}{0.pt}
	\centering
	\subfigure{
		\includegraphics[width=0.4\textwidth]{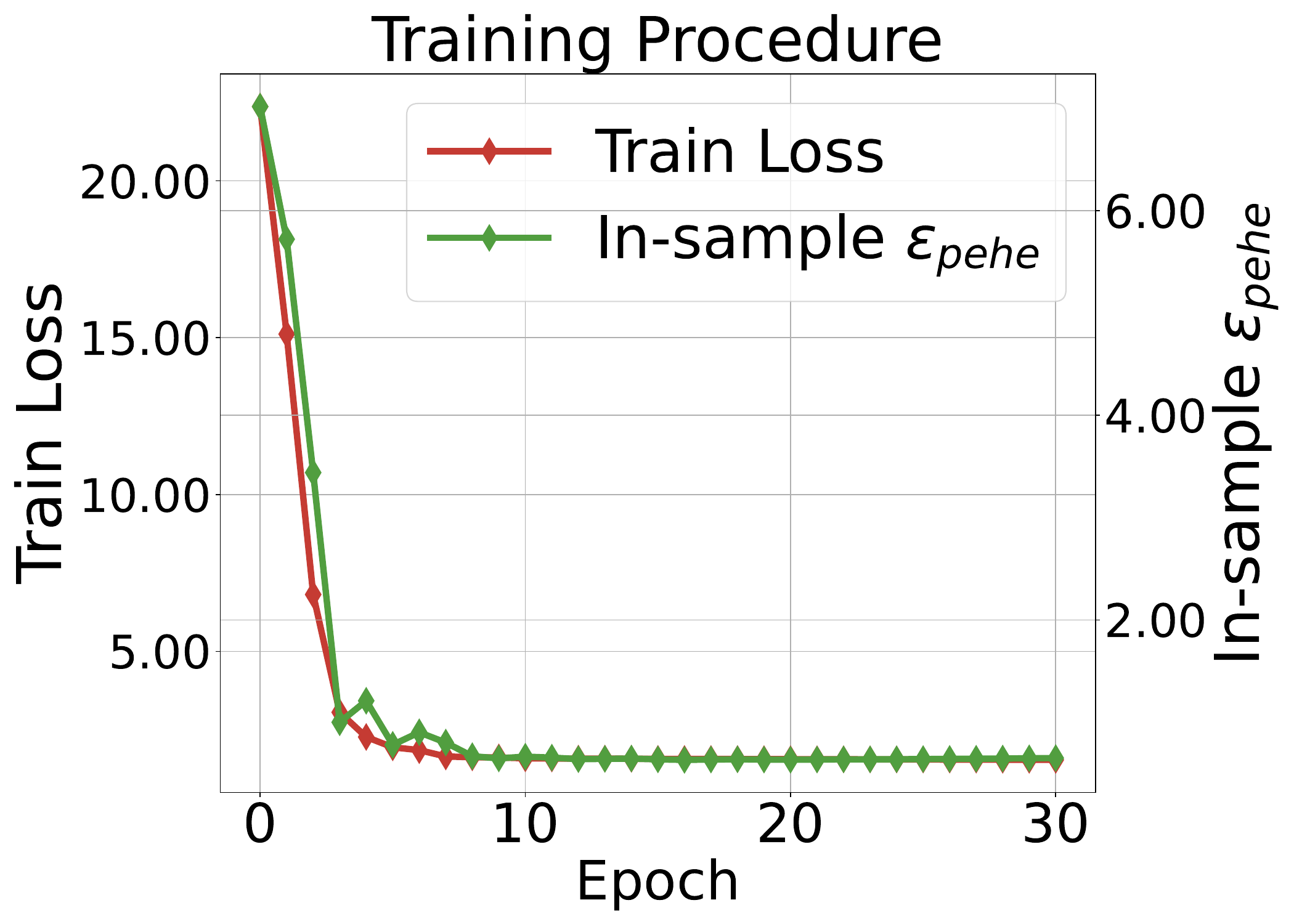}	
	}\qquad 
	\subfigure{
		\includegraphics[width=0.4\textwidth]{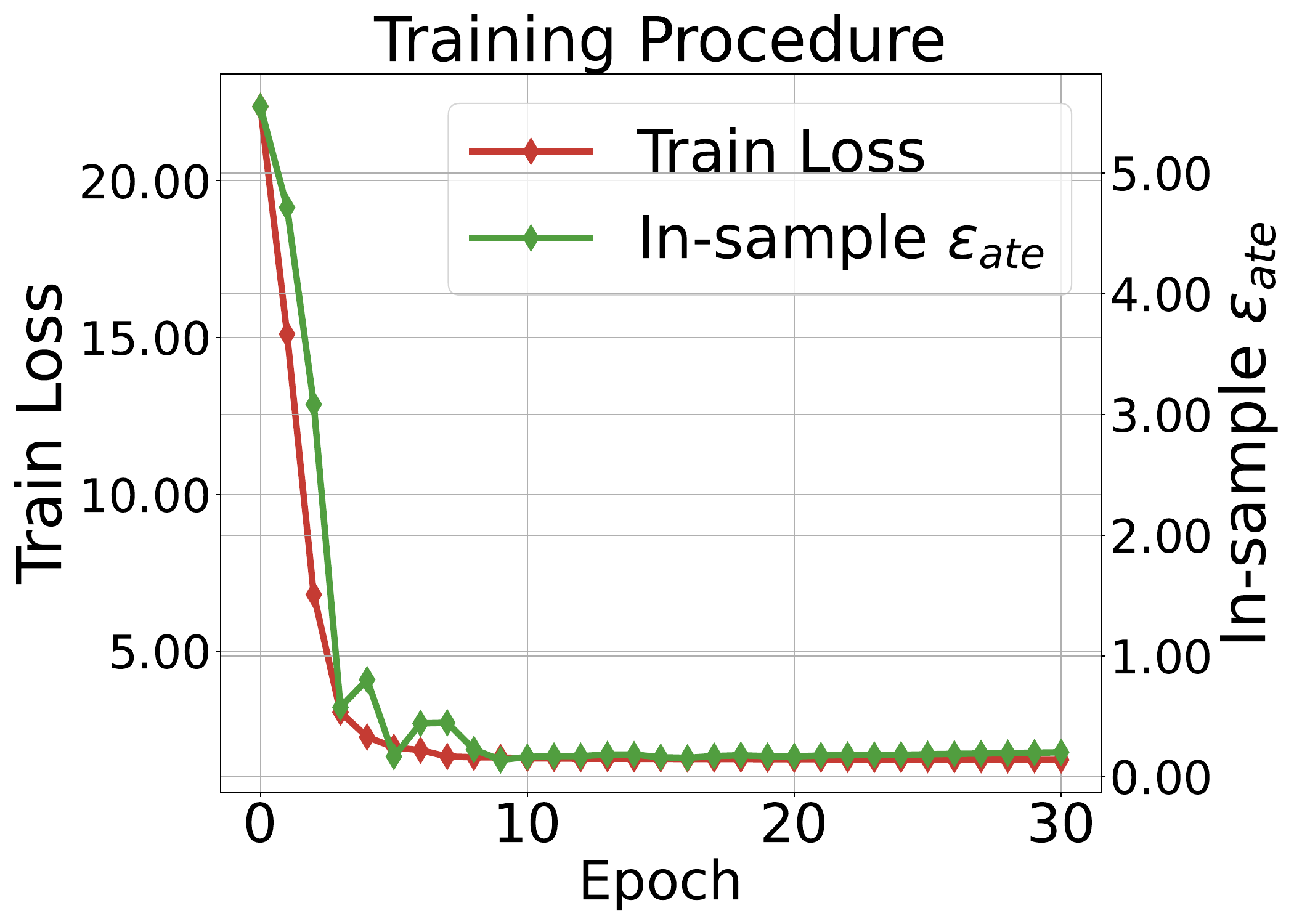}
	}\\
	\subfigure{
		\includegraphics[width=0.4\textwidth]{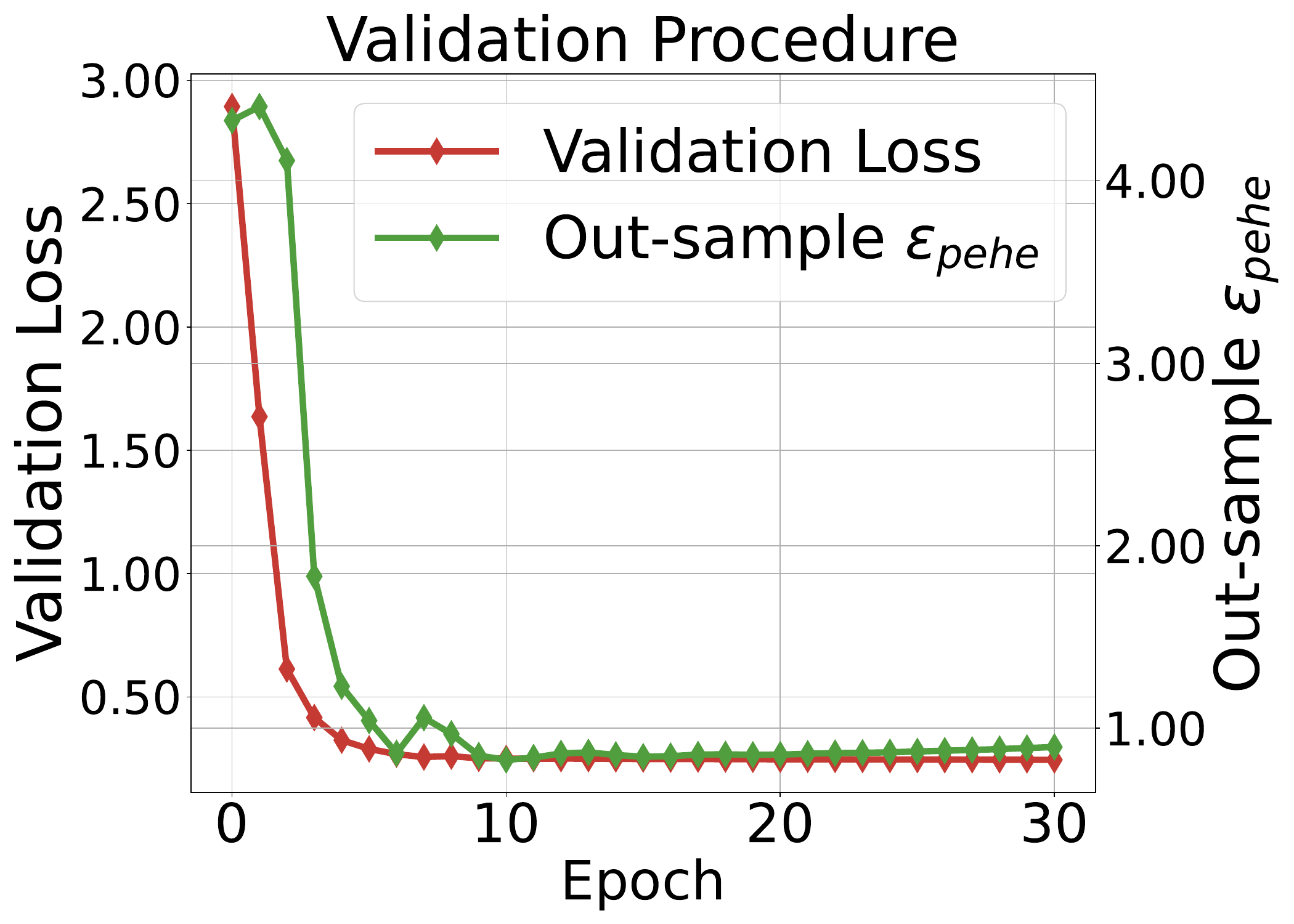}	\qquad 
	}
	\subfigure{
		\includegraphics[width=0.4\textwidth]{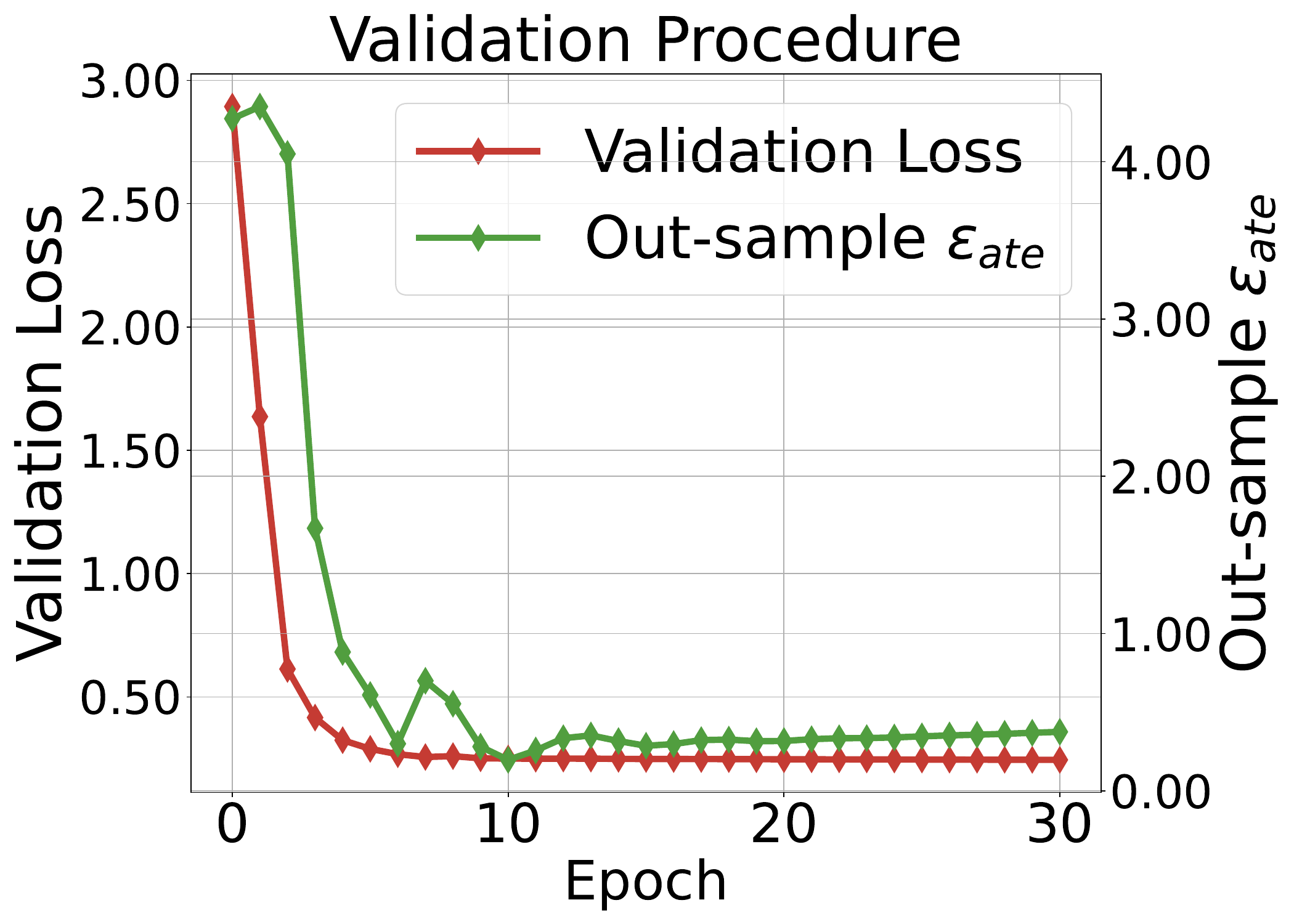}	
	}
	\vspace{-0cm}
	\caption{Visualizations on training procedure and validation procedure of GWIB on IHDP dataset.}\label{fig:loss}
	\vspace{-0cm}
\end{figure}

The algorithmic scheme of GWIB is outlined in Alg.~\ref{alg:main}. Each epoch involves a bi-level optimization process: first, we solve the optimal-transport based regularizer using the conditional gradient (CG) algorithm to iteratively obtain an optimal transport plan $T^*$.
Then, we update the model parameters $\{\phi, h_0, h_1\}$ using the classic stochastic gradient descent (SGD) method.

Specifically, in order to update the transport plan $\bm{T}$, the CG algorithm only requires leveraging the gradient $\nabla_{\bm{T}}\mathcal{L}$ computed in Section~\ref{sec-learn} and then adapting a
classic solver for Earth Mover's distance (EMD) such as a Lagrangian mass transport~\cite{bonneel2011displacement} algorithm to solve the optimal transport problem. 
Additionally, we propose a linesearch algorithm in Alg.~\ref{alg:ls} to search for the constrained
minimization of the second-degree regularization terms in Gromov-Wasserstein distances. 
In practice, we repeat the above procedure for adequate iterations to obtain an optimal transport plan.
Although the above quadratic problem can be provably non-convex, it can converge to a local stationary point using the CG method~\cite{lacoste2016convergence}.
Finally, given the optimal transport plan $\bm{T}^{(K)}$ after $K$ iterations, we then update the model parameters $\{\phi, h_0, h_1\}$ with classic SGD approach.

To evaluate the convergence of our proposed algorithms, we have visualized the training and validation results at each epoch. 
Figure~\ref{fig:loss} illustrates this analysis. 
It shows that the loss values for both training and validation decrease rapidly towards zero, indicating that our method achieves convergence swiftly. 
Concurrently, the performance in the ITE estimation task improves as the number of epochs increases, further demonstrating the effectiveness of our approach.

\section{Reproducibility}
\label{sec:repro}
In this section, we present additional details regarding the experiment setup and model implementation to ensure better reproducibility.
\subsection{Details of Experiment Setup}
\label{appendix-setup}
\subsubsection{Datasets}
We conduct experiments on two semi-synthetic datasets, namely IHDP and ACIC, to validate our proposed GWIB model.
The IHDP dataset~\cite{hill2011bayesian} focuses on the impact of special home visits for infants (treatment) on their future cognitive test scores (outcome). 
This dataset comprises 747 samples with 25-dimensional covariates collected from a real-world randomized experiment. 
Selection bias is manually introduced between the treatment and control groups by removing individuals with non-white mothers from the treated population.
The potential outcomes are generated using the NPCI package following~\cite{yao2018representation}.
The ACIC dataset~\cite{dorie2019automated} is obtained from the Collaborative Perinatal Project and serves as the dataset for the competition. 
Specifically, ACIC is a collection of semi-synthetic datasets whose covariates are taken from a large study conducted on pregnant women and their children to identifying causal factors leading to developmental disorders~\cite{niswander1972women}.
It includes data for 4802 patients with 58 covariates, encompassing both continuous and categorical variables.
For both datasets, we randomly shuffle them and allocate 63\% for training, 27\% for validation, and 10\% for testing.
\subsubsection{Baselines}
In our experiments, we compare various methods to demonstrate the effectiveness of our proposed GWIB, including 
$i$) Meta-learners: single network for outcome prediction without giving the treatment a special role (\textbf{S-Learner})~\cite{kunzel2019metalearners}, two separate networks for the treatment group and control group, respectively (\textbf{T-Learner})~\cite{kunzel2019metalearners};
$ii$) Matching based methods: k-nearest neighbors (\textbf{$k$-NN})~\cite{crump2008nonparametric}, propensity score match with treatment regression (\textbf{PSM})~\cite{caliendo2008some};
$iii$) Representation-based methods: two networks for predicting counterfactual outcomes with shared representation layer to alleviate the sample imbalance (\textbf{TARNet})~\cite{shalit2017estimating}, applying additional loss of discrepancy metric (e.g., Wasserstein distance and Maximum Mean Discrepancy distance) to balance the representation distributions between treatment group and control group (\textbf{CFR-Wass} and \textbf{CFR-MMD})~\cite{shalit2017estimating}, adjust treatment effect with propensity score layer and targeted regularization (\textbf{DragonNet})~\cite{shi2019adapting}, preservation of local similarity in balancing (\textbf{SITE})~\cite{yao2018representation}, combination of propensity score-based re-weighting and representation balancing (\textbf{BWCFR})~\cite{assaad2021counterfactual}, achieving outlier-free balance at mini-batch level with unbalanced optimal transport technique (\textbf{ESCFR})~\cite{wang2023optimal}.
Note that propensity scores in DragonNet, SITE, PSM, and BWCFR methods are estimated using logistic regression models following~\cite{caliendo2008some}.

\subsubsection{Training and Optimization}
\begin{table}[t]
    \label{tab:hyper}
    \setlength{\abovecaptionskip}{0.25cm}
    \setlength{\fboxrule}{0pt}
    \setlength{\fboxsep}{0.pt}
    \centering
    \caption{The configuration of hyperparameters in this paper.} 
        \begin{tabular}{c|c|c}  
            \hline \hline 
            Hyperparameter&Tuning range&Description \\  
            \hline
            lr & $\left[10^{-5},10^{-4},10^{-3},10^{-2},10^{-1}\right]$ & learning rate\\
            bs & $\left[16, 32, 64, 128\right]$ & batch size\\
            $\lambda$ & $\left[10^{-4},10^{-3},10^{-2},10^{-1}, 1\right]$ & weight of the proposed regularization in~\eqref{eq:reg}\\  
            $\beta$ & $\left[0.1, 0.3, 0.5, 0.7, 0.9\right]$ & weight inside $FGW_{2,\beta}(\bm{Z}_t,\bm{Z}_{1-t})$ in~\eqref{eq:fgwd}\\
            $d_{\phi}$ & $\left[16, 32, 64\right]$ & the dimension of hidden layer in $\phi$\\
            $d_{h}$ & $\left[16, 32, 64\right]$ & the dimension of hidden layer in $h_0$ and $h_1$\\
            \hline \hline
        \end{tabular}
\end{table}
In our experiments, we carefully tune all hyperparameters in GWIB using the validation set with grid search. 
First, we search for the optimal learning rate and batch size in the range of $\{10^{-5}, 10^{-4}, 10^{-3}, 10^{-2}, 10^{-1}\}$ and $\{16, 32, 64, 128\}$, respectively.
Then, we tune the hyperparameter $\lambda$ that trade-offs the balancing and outcome predictive loss over $\{10^{-4}, 10^{-3}, 10^{-2}, 10^{-1}, 1\}$.
The hyperparameter $\beta$ in fused Gromov-Wasserstein distance is tuned in the range of $\{0.1, 0.3, 0.5, 0.7, 0.9\}$.
The detailed configurations of hyperparameters are shown in Table~\ref{tab:hyper}.
The number of training epochs is set as 200.
For a fair comparison, all baselines are tuned within the same range as our proposed method. 
To prevent overfitting to the training set, we employ an early stopping mechanism with a patience of 30 epochs.
This precaution is particularly important as the ground-truth outcomes are generated in both IHDP and ACIC datasets, making them more prone to overfitting.
In practice, we implement our GWIB and all baselines using the PyTorch 1.10 framework with Adam as the optimizer.
All experiments are conducted on an NVIDIA A40 GPU and an Intel(R) Xeon(R) 5318Y Gold CPU @ 2.10GHz. 
More results of hyperparameter optimization can be referred to Appendix~\ref{appendix-results}.
\subsubsection{Evaluation}
For evaluating the performance of the treatment effect estimation task on IHDP and ACIC datasets, where both factual and counterfactual outcomes are generated from a known distribution, we utilize the absolute error in average treatment effect $\epsilon_{ATE}$, and the precision in estimation of heterogeneous effects $\epsilon_{PEHE}$ as metrics, which are denoted as:
\begin{eqnarray}
\begin{aligned}
	\epsilon_{ATE} &=  \frac{1}{|N|} \Bigl|\sum_{i=1}^{|N|} \left(y_i(x_i, 1)-y_i(x_i, 0)\right) - \left(\hat{y}_i(x_i, 1)-\hat{y}_i(x_i, 0)\right)\Bigr|\\
	\epsilon_{PEHE} &= \frac{1}{|N|} \sum_{i=1}^{|N|} \left(\left(y_i(x_i, 1)-y_i(x_i, 0)\right) - \left(\hat{y}_i(x_i, 1)-\hat{y}_i(x_i, 0)\right)\right)^2,
\end{aligned}
\end{eqnarray}
where $N = N_0 + N_1$ is the total number of individuals in control and treatment groups, $\hat{y}_i(x_i, 1)$ and $\hat{y}_i(x_i, 0)$ are the estimated potential outcomes while $y_i(x_i, 1)$ and $y_i(x_i, 0)$ are the ground-truth ones.
To evaluate the generalizability of our proposed GWIB model, we have conducted two types of experiments, namely, in-sample experiments and out-sample experiments.
In the in-sample scenario, the testing samples are drawn from the training distribution, while in the out-sample scenario, the testing samples are drawn from distributions outside the training set. 
In practice, we run each experiment five times by varying the random seeds and report the mean and standard deviation of the performances.

\subsection{Details of Model Implementation}
\subsubsection{Details of Prediction Model}
As aforementioned, our method is based on the backbone consisting of the representation mapping $\phi$ as well as the regression estimators $h_0, h_1$ for control and treatment groups, respectively, which are all parameterized by neural networks.
In specific, the backbone includes the following components:
\begin{itemize}
\item A two-layer multilayer perceptron (MLP) with dropout and ELU activation function for representation mapping $\phi$.
\item Two single-layer multilayer perceptron (MLP) with dropout and ELU activation function for regression estimators $h_0$ and $h_1$, respectively.
\end{itemize}
The above prediction model utilizes MSE loss for supervised training over observational data $\{x_m,y_m(x_m,0)\}_{m=1}^{N_0}$ and $\{x_n,y_n(x_n,1)\}_{n=1}^{N_1}$.
Unless otherwise specified, the dimensions for the two-layer MLP in $\phi$ are set as $\left[32, 16\right]$, while that for the single MLP in each head $h_0$ and $h_1$ is set as 16 by default.
We set the dropout rate as 0.1 in all experiments.
Actually, the prediction model corresponds to the representation-based method TARNet~\cite{shalit2017estimating}, where we can observe in Table~\ref{exp-main} that it demonstrates poorer performance than most of the other methods, since it does not explicitly add any regularization to address the selection bias.

\subsubsection{Details of GWIB}
GWIB method is designed in a general way, which can be adaptable to different foundation model, including the aforementioned two-head prediction model adopted in this paper.
In practice, our implementation is built upon the Python Optimal Transport (POT 0.9.0)  package~\cite{flamary2021pot}.
However, in the original POT, there is no available API function to fuse more than two transport plans, where we need to fuse at least three transport plans for our proposed regularizer in~\eqref{eq:gwib2}.
Therefore, we manually extend the original API of fused Gromov-Wasserstein distance function based on a novel conditional gradient (CG) algorithm and a specifically designed line search (LS) approach, where we detail them in Algs.~\ref{alg:main} and~\ref{alg:ls}, respectively.  
\section{More Experimental Results}
\label{appendix-results}
\subsection{Sensitive Analysis of Hyperparameter $\lambda$}
\begin{figure}[t]
\setlength{\abovecaptionskip}{0.2cm}
\setlength{\fboxrule}{0.pt}
\setlength{\fboxsep}{0.pt}
\centering
\subfigure{
	\includegraphics[width=0.4\textwidth]{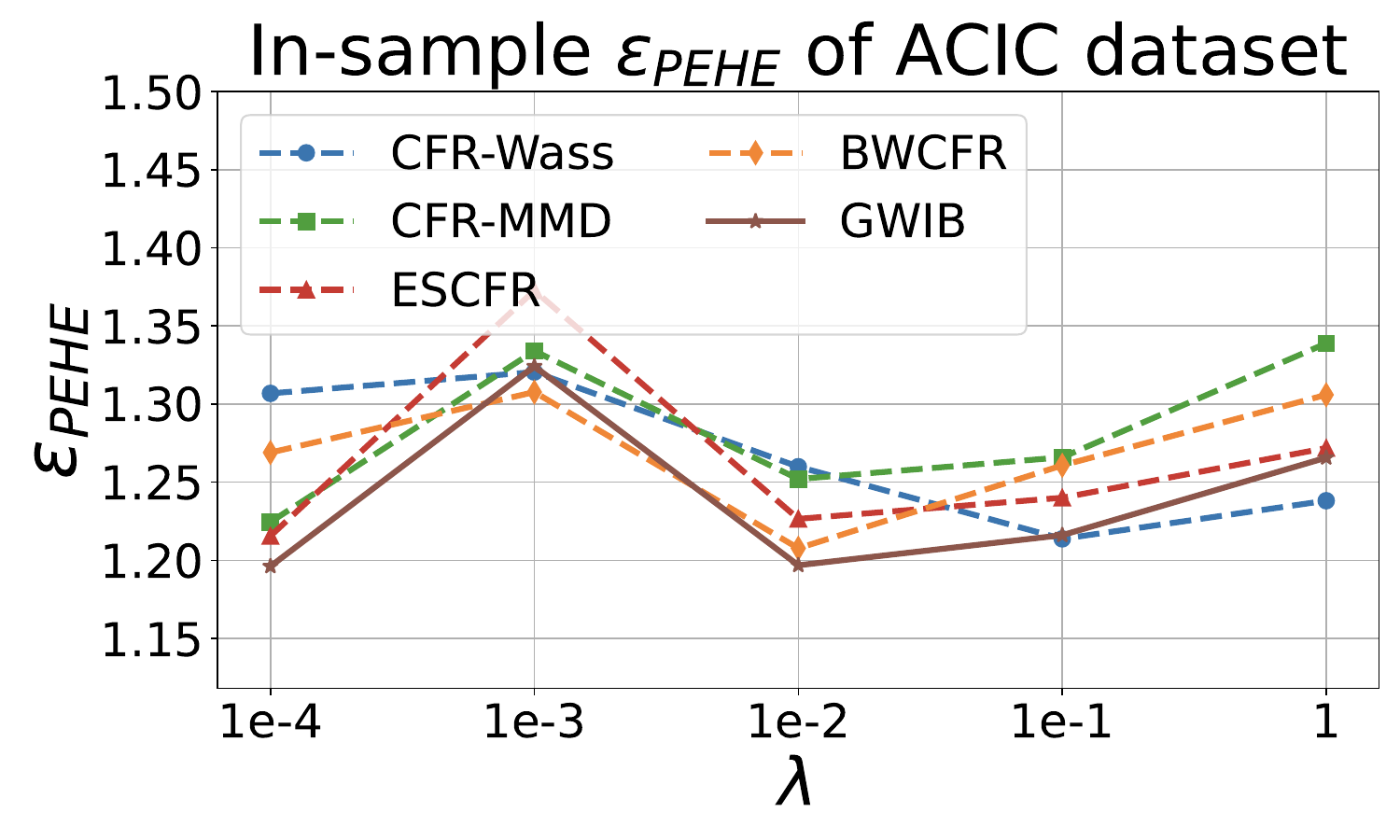}	
}\qquad
\subfigure{
	\includegraphics[width=0.4\textwidth]{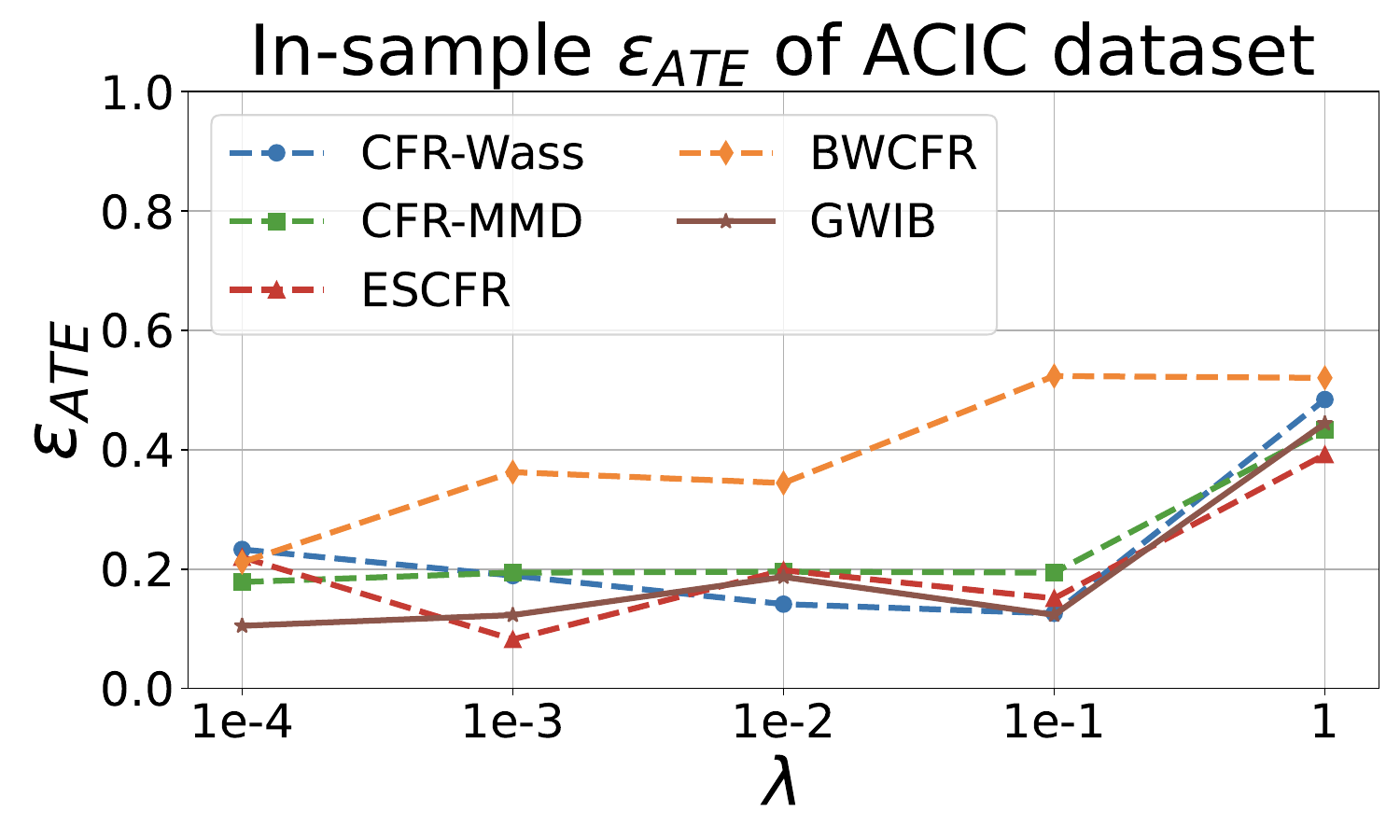}
}\\
\subfigure{
	\includegraphics[width=0.4\textwidth]{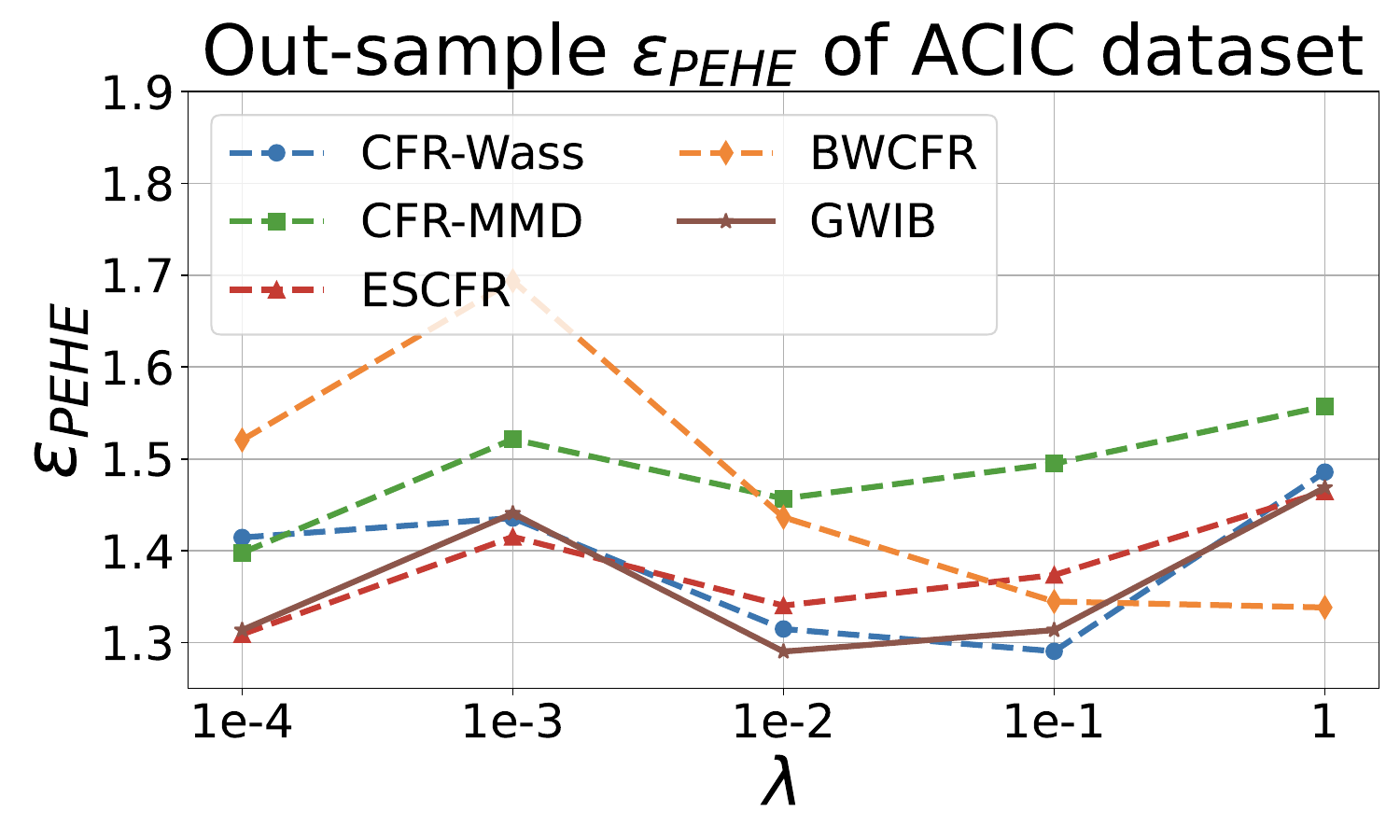}	
}\qquad
\subfigure{
	\includegraphics[width=0.4\textwidth]{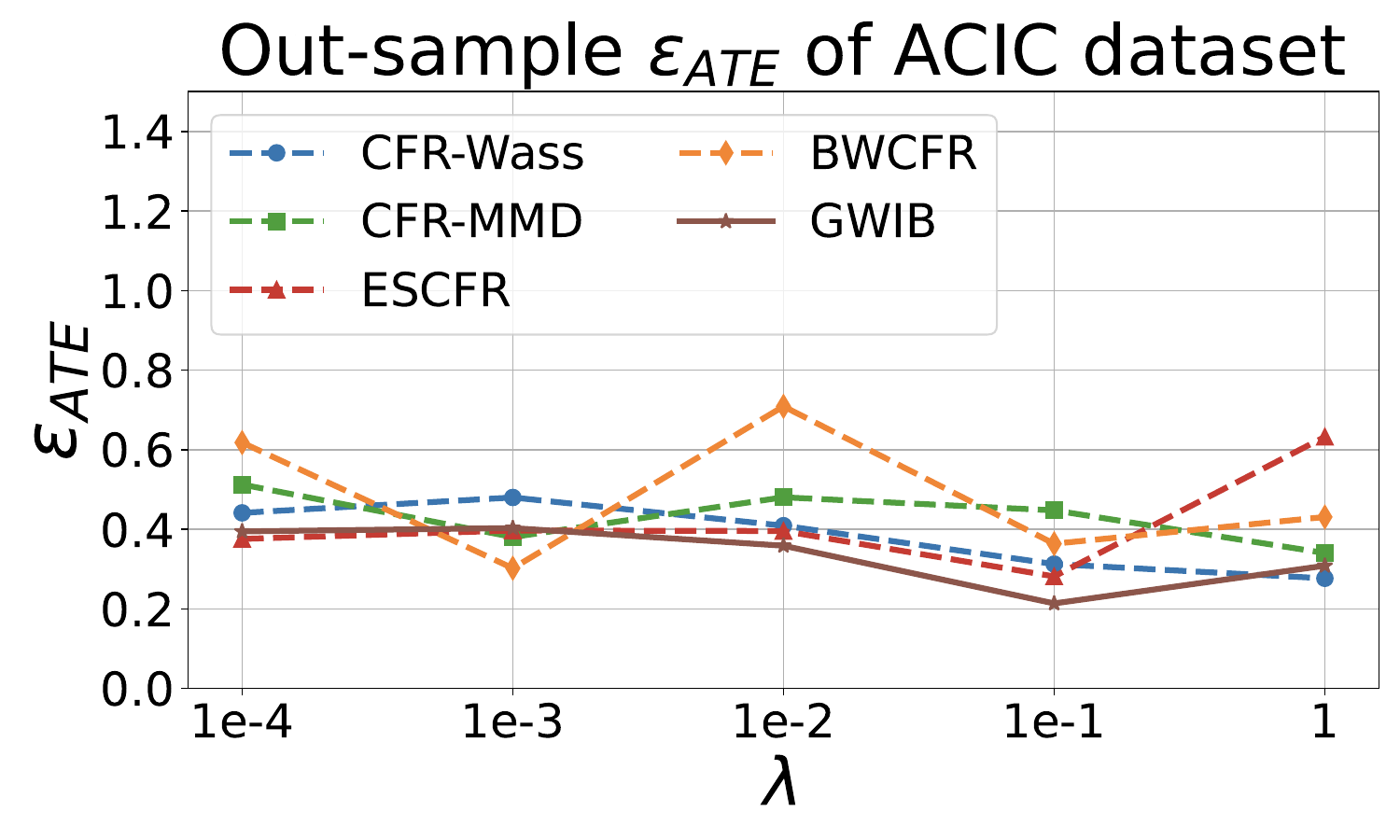}	
}
\caption{$\epsilon_{PEHE}$ and $\epsilon_{ATE}$ of different values of $\lambda$ in both in-sample and out-sample experiments on ACIC dataset.
}\label{fig-lambda-ACIC}
\vspace{-0cm}
\end{figure}

\begin{figure}[t]
\setlength{\abovecaptionskip}{0.2cm}
\setlength{\fboxrule}{0.pt}
\setlength{\fboxsep}{0.pt}
\centering
\subfigure{
	\includegraphics[width=0.4\textwidth]{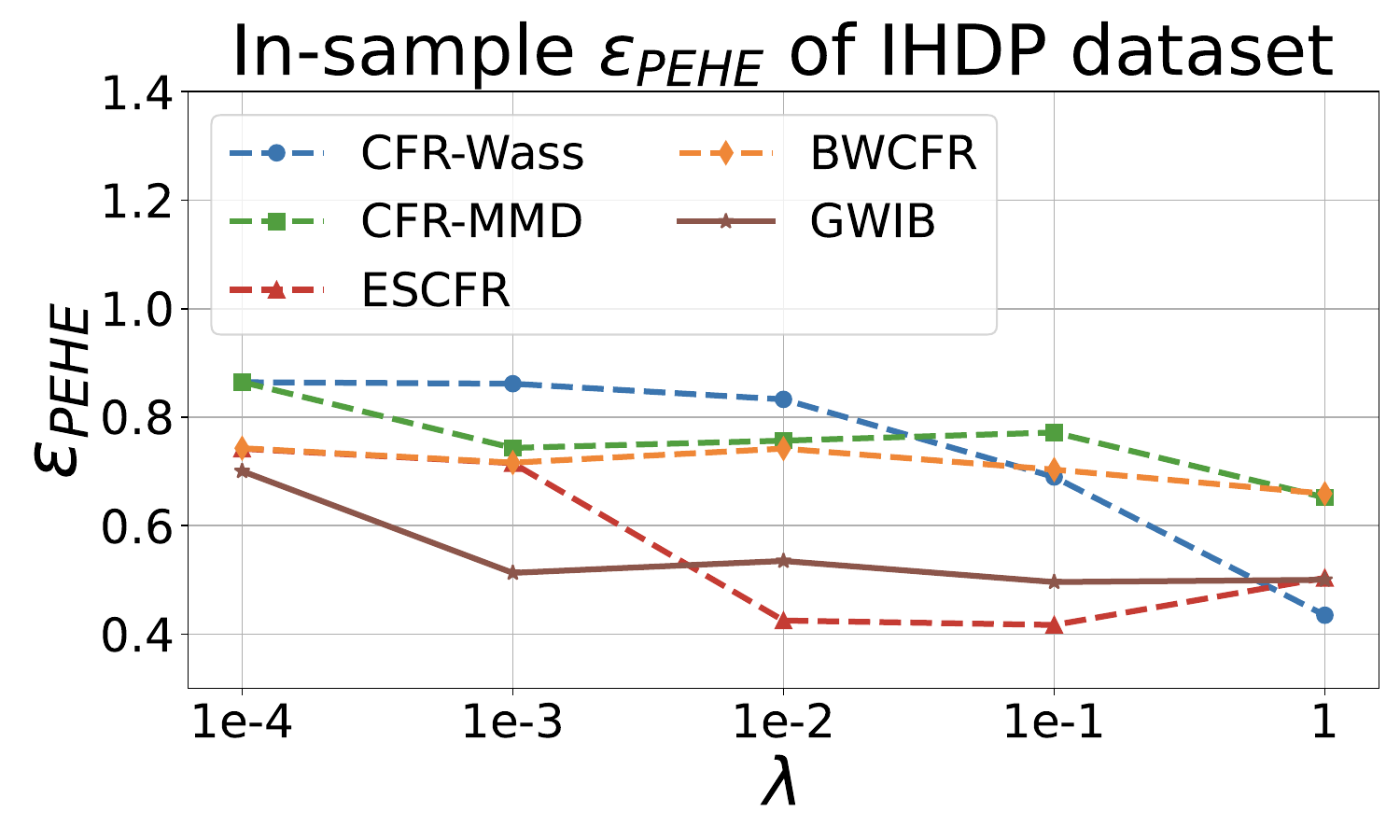}	
}\qquad
\subfigure{
	\includegraphics[width=0.4\textwidth]{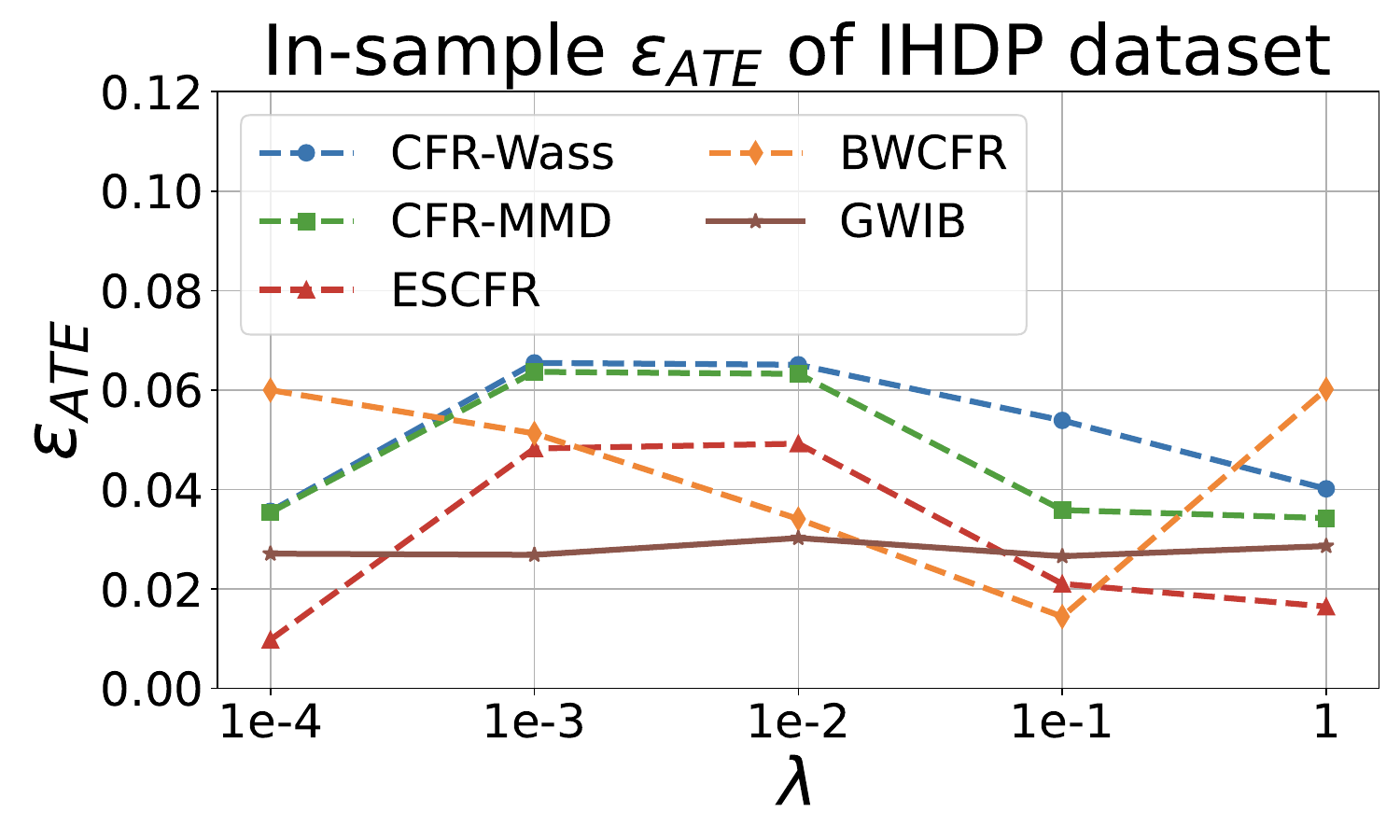}
}
\subfigure{
	\includegraphics[width=0.4\textwidth]{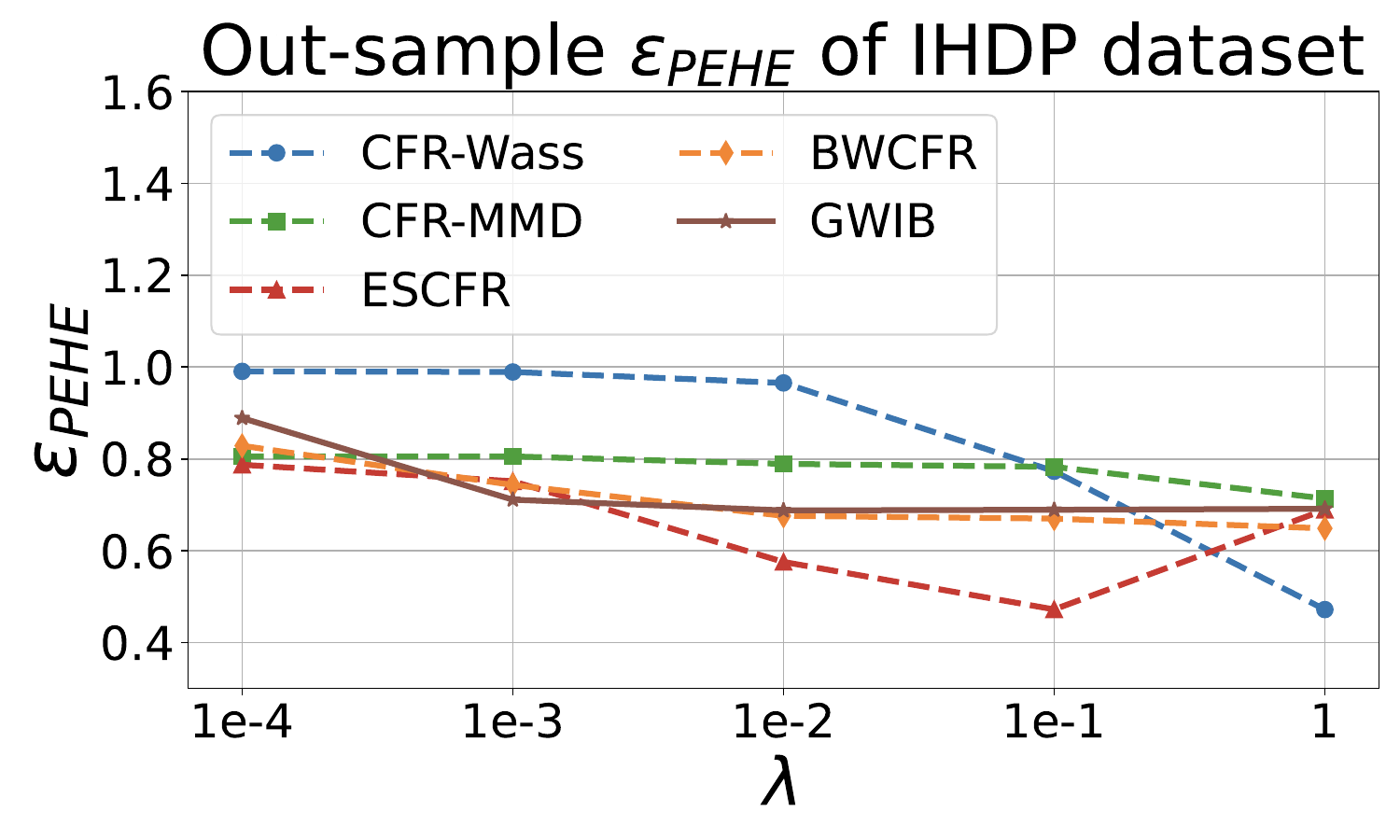}	
}\qquad
\subfigure{
	\includegraphics[width=0.4\textwidth]{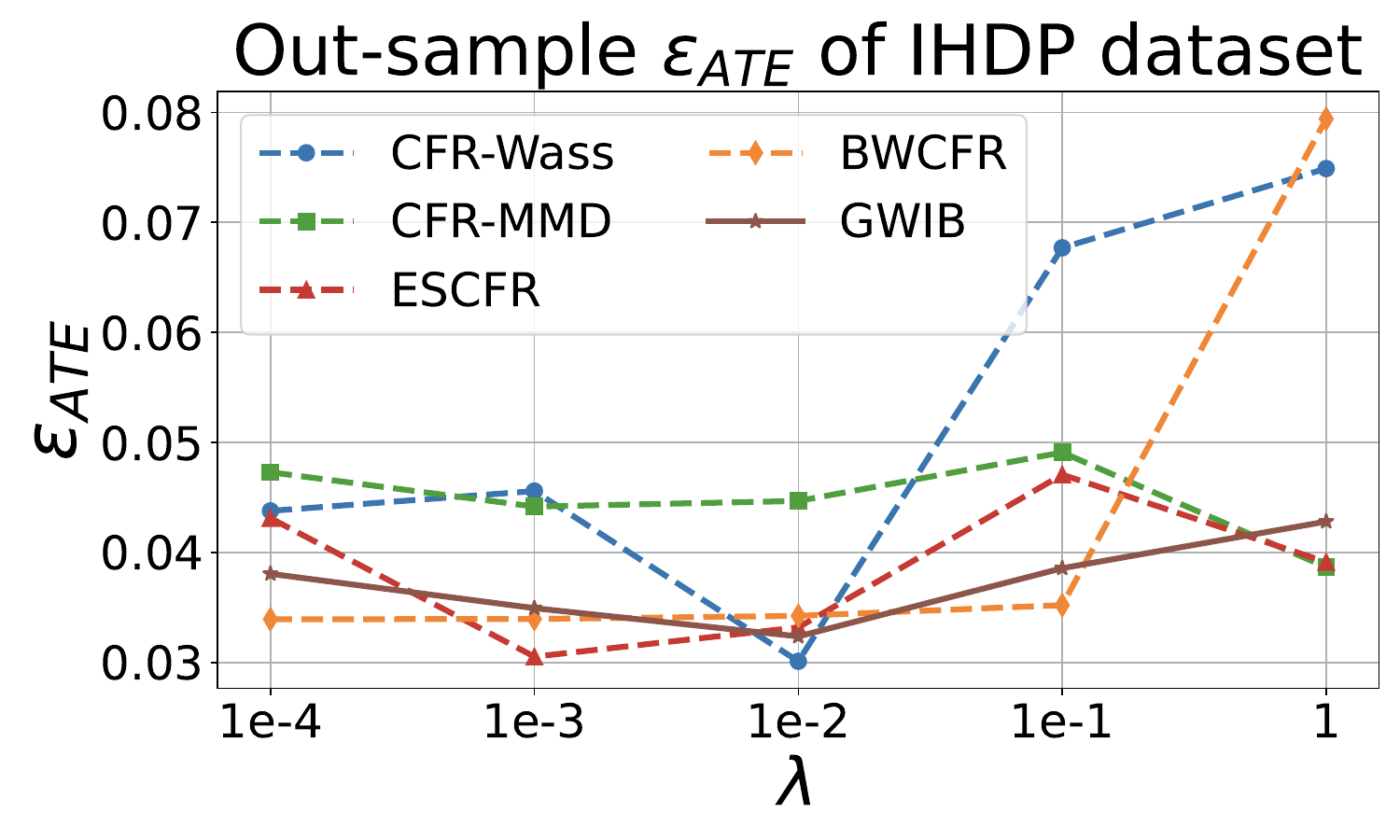}	
}
\vspace{-0cm}
\caption{$\epsilon_{PEHE}$ and $\epsilon_{ATE}$ of different values of $\lambda$ in both in-sample and out-sample experiments on IHDP dataset.}\label{fig-lambda-IHDP}
\vspace{-0cm}
\end{figure}
According to~\cite{shalit2017estimating}, the hyperparameter $\lambda$ in~\eqref{eq:classic_cfr} is crucial in determining the trade-off between the outcome prediction task and the representation balancing task for CFR-based models, including CFR-Wass, CFR-MMD, ESCFR, BWCFR, and GWIB methods in our paper. 
A higher value of $\lambda$ often indicates more balanced representations in the latent space. 
Therefore, it is necessary to verify the effectiveness of our proposed GWIB method under different choices of $\lambda$. 
Similar to the main experiments, we conduct both in-sample and out-sample experiments and report $\epsilon_{ATE}$ and $\epsilon_{PEHE}$ as metrics.
The hyperparameter $\lambda$ is varied within the range of $\{10^{-4}, 10^{-3}, 10^{-2}, 10^{-1}, 1\}$. 
As demonstrated in Figure~\ref{fig-lambda-ACIC} and~\ref{fig-lambda-IHDP}, we make the following observations:

\begin{enumerate}
\item The impact of the hyperparameter $\lambda$ on the performance of the treatment effect estimation task is evident. 
As \(\lambda\) increases, we observe subtle differences in the tendencies of \(\epsilon_{PEHE}\) and \(\epsilon_{ATE}\) across all methods. 
Specifically, the best performance for $\epsilon_{ATE}$ is achieved at a moderate value of $\lambda$, while $\epsilon_{PEHE}$ often yields higher performances with extreme higher or lower values of $\lambda$. 
This is not surprising, since $\epsilon_{PEHE}$ assesses precision in predicting treatment effects at the individual level by averaging squared differences, while $\epsilon_{ATE}$ evaluates overall accuracy by averaging absolute differences in predicted and observed average treatment effects across the entire population.
\item In most cases, CFR-Wass and CFR-MMD methods, which employ pure representation balancing with different discrepancy metrics, exhibit the worst performance in both $\epsilon_{PEHE}$ and $\epsilon_{ATE}$ metrics. 
Besides, ESCFR method has typically obtained the sub-optimal performances across different values of $\lambda$, which introduces the unbalanced optimal transport (UOT) technique to relax marginal constraints for flexible balancing at the mini-batch level.
However, all of the above baselines have overlooked the over-enforcing balance issue that would obtain undesired trivial distributions in latent space, which may inadvertently remove information that is predictive of outcomes, thus leading to deteriorated performances in the treatment effect estimation task.
Although BWCFR can address the over-enforcing balance issue by reweighting the covariate a priori, its performance is significantly dependent on the accurate estimation of propensity scores, which is known to be challenging and often demonstrates high variance~\cite{caliendo2008some}.

\item Despite being highly sensitive to the value of $\lambda$, our proposed GWIB consistently outperforms other CFR-based baselines in most cases, demonstrating the superiority of the proposed Gromov-Wasserstein information bottleneck-based regularizer. By approximating the neural network $\phi$ as a Monge map, we are able to effectively preserve the structural information of the latent representations, stemming from their original covariate distributions, while simultaneously achieving a compression of their complexities. Additionally, the regularizer has realigned the cross-group individual-level correspondence between control and treatment groups in a reasonable manner, further improving the precision of counterfactual outcome prediction.
\end{enumerate}

\subsection{Sensitive Analysis of Hyperparameter $\beta$}
\begin{table*}[t]
\caption{Sensitive analysis of hyperparameter $\beta$ on ACIC and IHDP datasets.}
\label{exp:beta}
\LARGE
{
\resizebox{\textwidth}{!}{
    \begin{tabular}{c|cccc|cccc}		
        \hline 
        \hline
        Datasets&\multicolumn{4}{c|}{ACIC}&\multicolumn{4}{c}{IHDP} \\ \midrule
        Test Types&\multicolumn{2}{c}{In-sample}&\multicolumn{2}{c|}{Out-sample}&\multicolumn{2}{c}{In-sample}&\multicolumn{2}{c}{Out-sample} \\ \midrule
        Methods & $\epsilon_{ATE}$ & $\epsilon_{PEHE}$  & $\epsilon_{ATE}$ & $\epsilon_{PEHE}$& $\epsilon_{ATE}$ & $\epsilon_{PEHE}$& $\epsilon_{ATE}$ & $\epsilon_{PEHE}$ \\ \midrule
        $\beta=\text{0.1}$ & 0.3607$_{\pm \text{0.0415}}$ & 1.3219$_{\pm \text{0.0081}}$ & 0.7454$_{\pm \text{0.2248}}$ & 1.5727$_{\pm \text{0.0496}}$ & 0.1112$_{\pm \text{0.0015}}$ & 0.7612$_{\pm \text{0.0017}}$ & 0.2485$_{\pm \text{0.0051}}$ & 0.9799$_{\pm \text{0.0025}}$ \\ 
        $\beta=\text{0.3}$ & 0.3236$_{\pm \text{0.0810}}$ & 1.3447$_{\pm \text{0.0074}}$ & 0.7042$_{\pm \text{0.0786}}$ & 1.5378$_{\pm \text{0.0061}}$ & 0.5165$_{\pm \text{0.0131}}$ & 1.0503$_{\pm \text{0.0257}}$ & 0.7804$_{\pm \text{0.0100}}$ & 1.3053$_{\pm \text{0.0434}}$ \\ 
        $\beta=\text{0.5}$ & 0.3041$_{\pm \text{0.0148}}$ & 1.3373$_{\pm \text{0.0123}}$ & 0.7289$_{\pm \text{0.1342}}$ & 1.5233$_{\pm \text{0.0841}}$ & 0.6293$_{\pm \text{0.0036}}$ & 1.1299$_{\pm \text{0.0066}}$ & 0.9600$_{\pm \text{0.0024}}$ & 1.3667$_{\pm \text{0.0027}}$ \\ 
        $\beta=\text{0.7}$ & 0.3389$_{\pm \text{0.0310}}$ & 1.3668$_{\pm \text{0.0116}}$ & 0.5794$_{\pm \text{0.0388}}$ & 1.4795$_{\pm \text{0.0202}}$ & 0.6734$_{\pm \text{0.0072}}$ & 1.1555$_{\pm \text{0.0129}}$ & 0.9979$_{\pm \text{0.0058}}$ & 1.4051$_{\pm \text{0.0036}}$ \\ 
        $\beta=\text{0.9}$ & 0.3830$_{\pm \text{0.0993}}$ & 1.3781$_{\pm \text{0.0256}}$ & 0.7384$_{\pm \text{0.2047}}$ & 1.6156$_{\pm \text{0.0575}}$ & 0.6589$_{\pm \text{0.0146}}$ & 1.1516$_{\pm \text{0.0213}}$ & 0.9991$_{\pm \text{0.0136}}$ & 1.3859$_{\pm \text{0.0119}}$ \\ 
        \hline \hline
    \end{tabular}
    }
}
\end{table*}

In~\eqref{eq:fgwd}, we have introduced a hyperparameter $\beta$ in the fused Gromov-Wasserstein distance $FGW_{2,\beta}(\bm{Z}_t, \bm{Z}_{1-t})$, to control the relative weight between Gromov-Wasserstein $GW_2(\bm{Z}_t, \bm{Z}_{1-t})$ and Wasserstein distances $W_2(\bm{Z}_t, \bm{Z}_{1-t})$. 
Since the mechanism by which $\beta$ influence the final treatment effect estimation task is not known a priori, a grid search is conducted to find the optimal value of $\beta$ in the range of $\{0.1, 0.3, 0.5, 0.7, 0.9\}$ on ACIC and IHDP datasets. 
The results, as shown in Table~\ref{exp:beta}, reveal the following observations:
\begin{enumerate}
\item By adjusting the value of $\beta$, our GWIB method exhibits superior stability in terms of $\epsilon_{ATE}$ metric when compared to $\epsilon_{PEHE}$ across both datasets, particularly in out-of-distribution experiments. Therefore, in practical scenarios such as advertising systems, it is advisable to fine-tune our GWIB method according to the $\epsilon_{ATE}$ values to guarantee more consistent performance during online deployments.

\item Furthermore, the performance variations are notably distinct between the ACIC and IHDP datasets. 
Specifically, on the ACIC dataset, optimal performance is attained with a moderate value of $\beta$, whereas on the IHDP dataset, a smaller $\beta$ value yields superior results. 
We speculate that the data in the IHDP dataset possesses a more geometric structure compared to the ACIC dataset, and thus, a higher weighting of $GW_2(\bm{Z}_t, \bm{Z}_{1-t})$ aids in capturing this inherent structure. 
Therefore, it emphasizes the need to independently fine-tune the value of $\beta$ in different applications to obtain the optimal performance.
\end{enumerate}

\subsection{Additional Visualizations on Balancing Representations}
\begin{figure}[t]
\setlength{\abovecaptionskip}{0.2cm}
\setlength{\fboxrule}{0.pt}
\setlength{\fboxsep}{0.pt}
\centering
\subfigure[$\lambda = 10^{-4}$]{
	\includegraphics[width=0.23\textwidth]{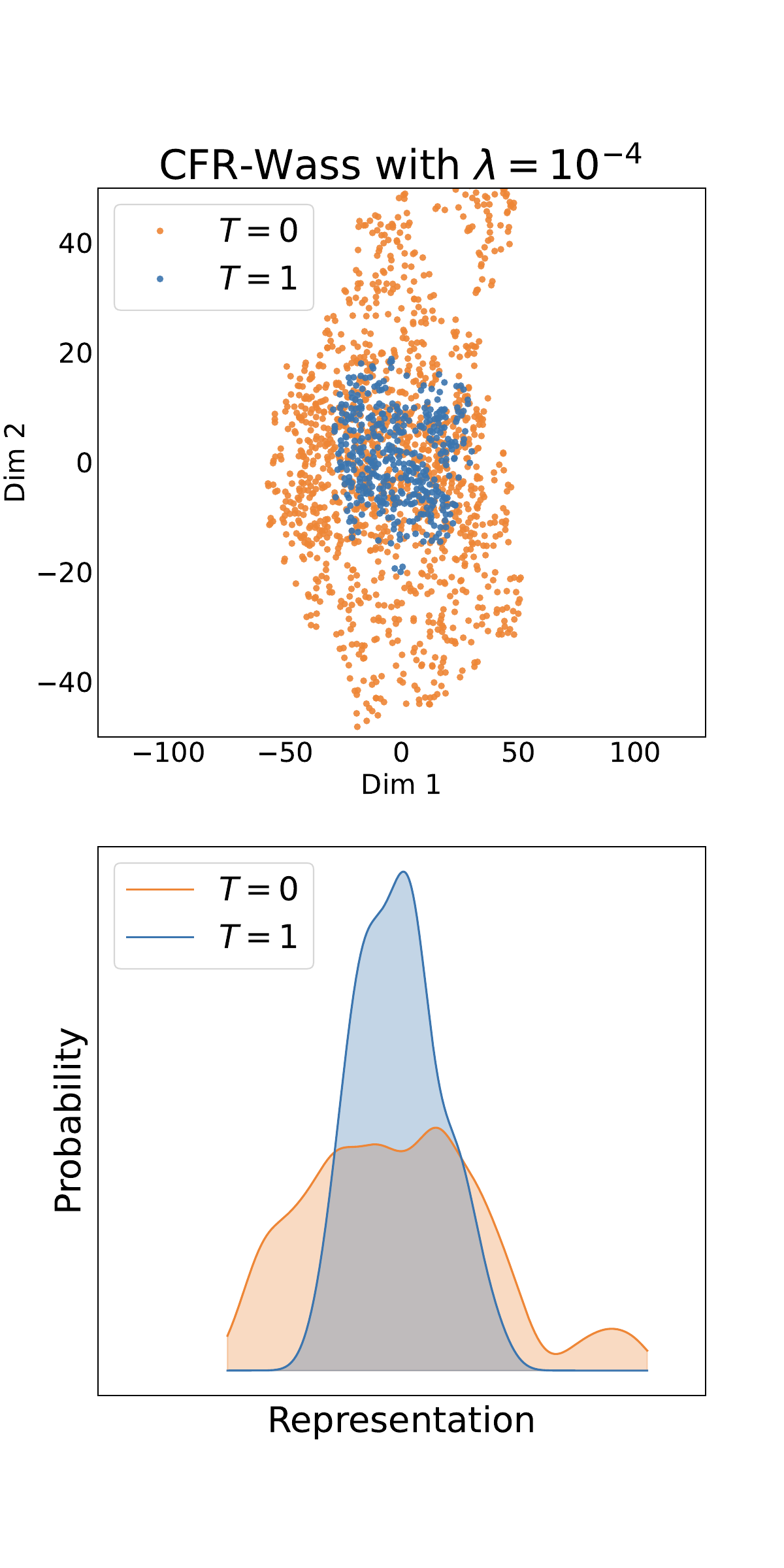}	
}
\hspace{-0.2cm}
\subfigure[$\lambda = 10^{-4}$]{
	\includegraphics[width=0.23\textwidth]{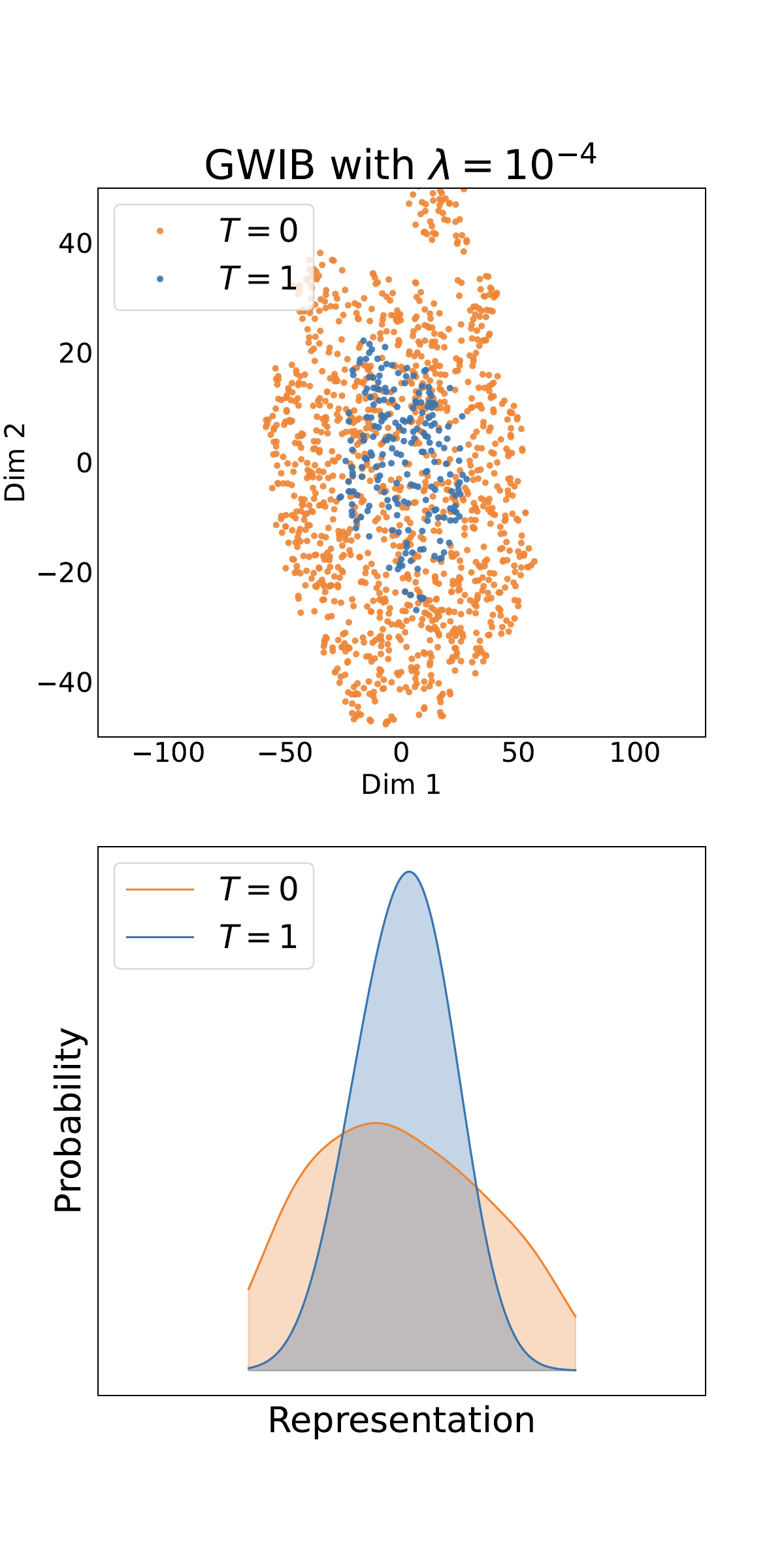}
}
\hspace{-0.2cm}
\subfigure[$\lambda = 10^{-2}$]{
	\includegraphics[width=0.23\textwidth]{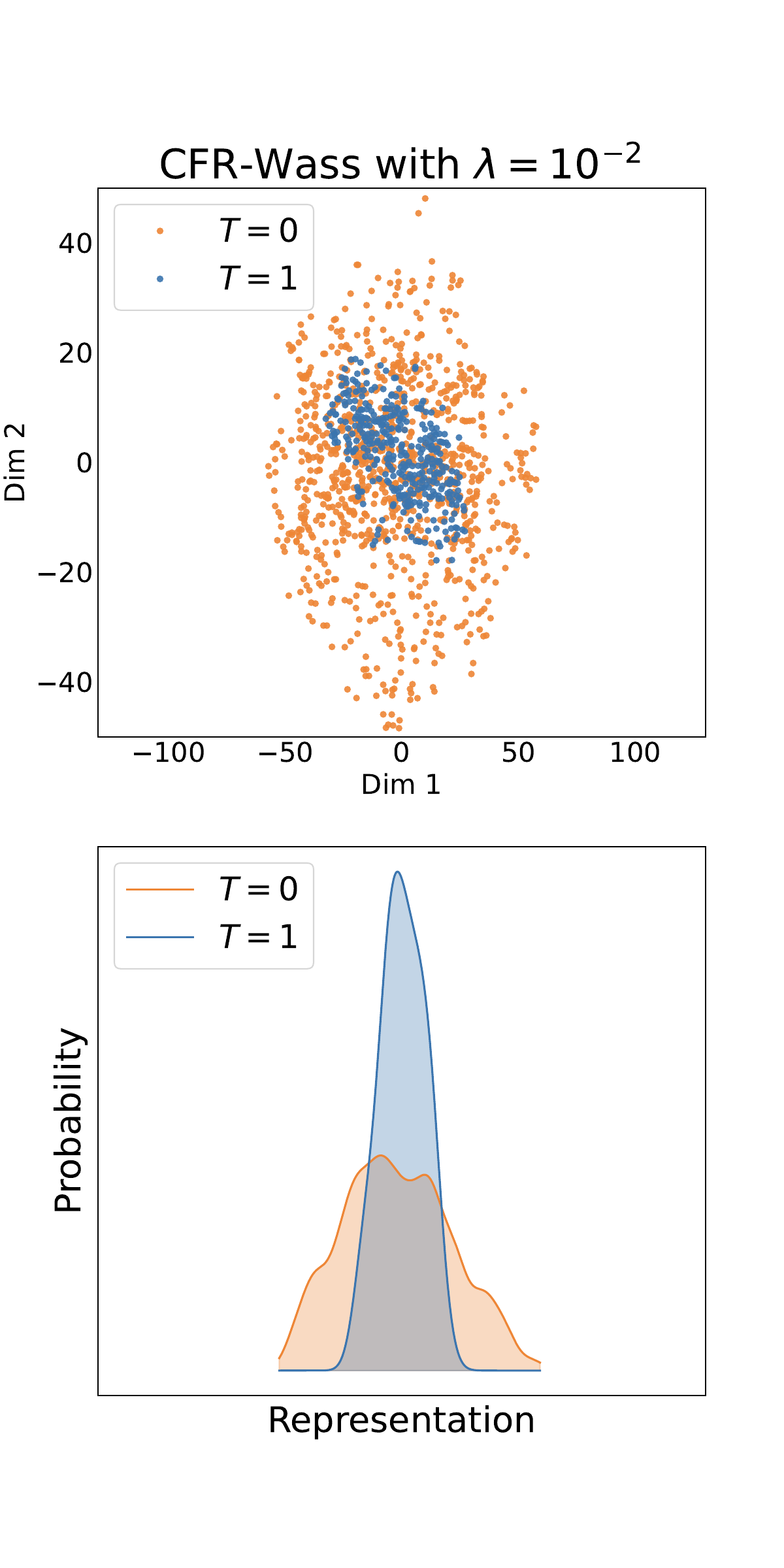}	
}
\hspace{-0.2cm}
\subfigure[$\lambda = 10^{-2}$]{
	\includegraphics[width=0.23\textwidth]{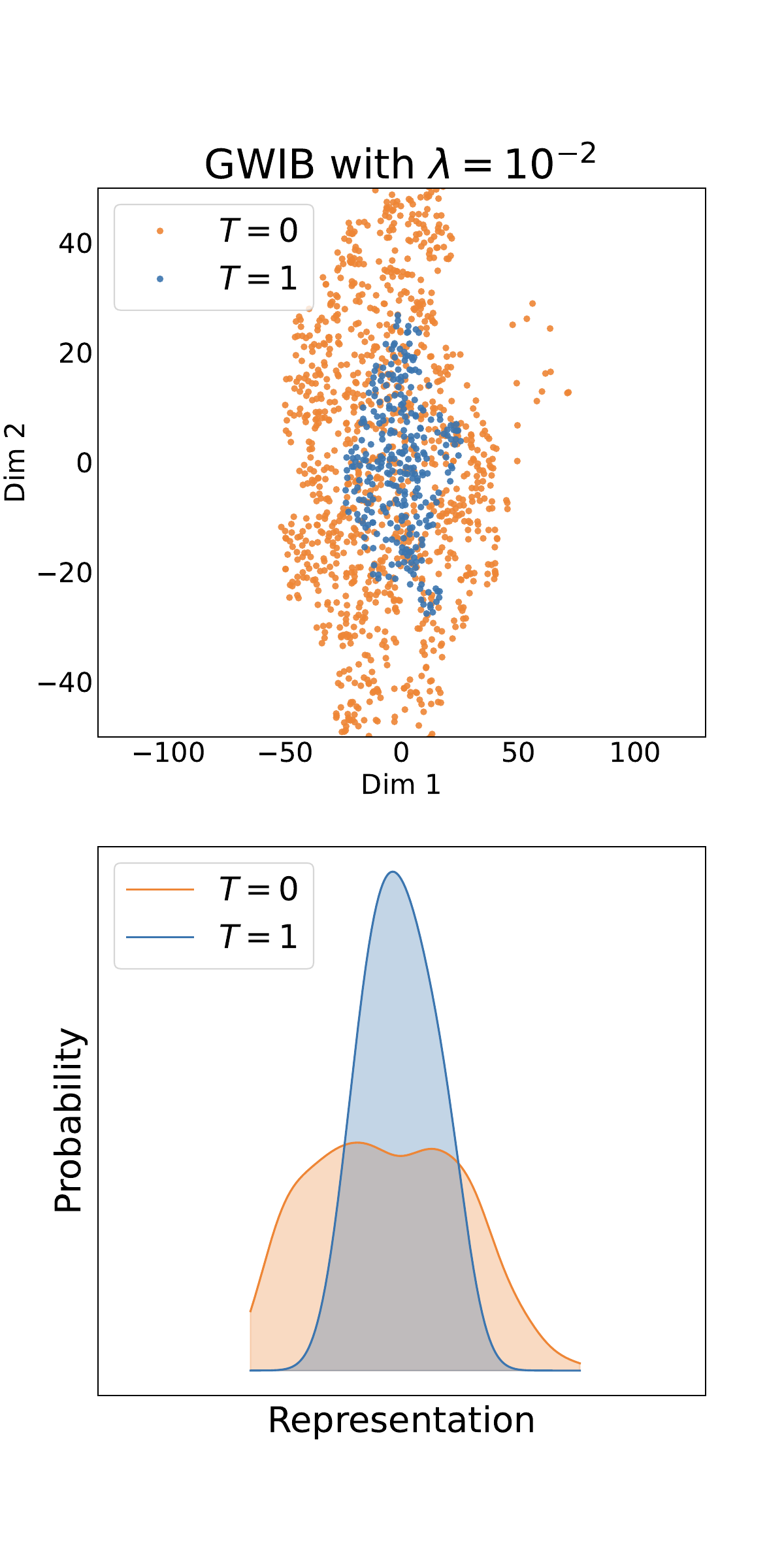}	
}
\vspace{-0cm}
\caption{Visualizations of the representation distributions generated from CFR-Wass and GWIB methods, respectively. The figures in the upper row represent scatter plots with t-SNE visualization, and the figures in the lower row show their corresponding distributions. The orange points or distributions represent the individuals in the control group, while the blue points or distributions indicate individuals in the treatment group.
}\label{fig:lambda_tsne}
\vspace{-0cm}
\end{figure}
In this section, we expand our experiment to visualize the sample distributions of the control group ($T=0$) and treatment group ($T=1$) in the latent space. 
Specifically, we compare the visual outcomes between the CFR-Wass and GWIB methods on ACIC dataset, varying the hyperparameter $\lambda$ within $\{10^{-4}, 10^{-2}\}$, which governs the strength of their corresponding balance regularizers. 
For each experiment, we present a scatter plot in the upper row, akin to the experiments in Section~\ref{sec:exp-ite}, accompanied by an additional probabilistic density function in the lower row for enhanced clarity. 

As depicted in Figure~\ref{fig:lambda_tsne}, we readily observe that both CFR-Wass and GWIB methods effectively balance the samples in the latent space, with highly overlapping covariates between the treatment and control groups. 
However, without additional approximate regularization, CFR-Wass may yield trivial latent distributions where all samples are projected with similar representations, thereby potentially undermining the predictive capability for ITE estimation tasks. 
This issue is exacerbated with larger values of $\lambda$ (e.g., the results in the third column compared to the first column in Figure~\ref{fig:lambda_tsne}). 
Fortunately, GWIB successfully addresses this over-enforcing balance issue by guiding the mapping network $\phi$ to approximate a Monge map. 
This approach naturally preserves information from the original covariate distributions while minimizing distribution discrepancies in the latent space.

\end{document}